\documentclass[10pt,twocolumn,letterpaper]{article}

\usepackage{cvpr}              

\usepackage{graphicx}
\usepackage{amsmath}
\usepackage{amssymb}
\usepackage{booktabs}
\usepackage{epsfig}
\usepackage{multirow}
\usepackage{pifont}
\newcommand{\xmark}{\ding{55}}%
\usepackage{color,colortbl}
\usepackage{xcolor}

\usepackage[accsupp]{axessibility}

\definecolor{Gray}{gray}{0.9}
\definecolor{Gray2}{gray}{0.95}

\usepackage[pagebackref,breaklinks,colorlinks]{hyperref}

\usepackage[capitalize]{cleveref}
\crefname{section}{Sec.}{Secs.}
\Crefname{section}{Section}{Sections}
\Crefname{table}{Table}{Tables}
\crefname{table}{Tab.}{Tabs.}

\def\confName{CVPR}
\def\confYear{2022}

\begin{document}
\title{RelTransformer: A Transformer-Based Long-Tail Visual Relationship Recognition}

\author{
  Jun Chen\textsuperscript{\rm 1},   Aniket Agarwal\textsuperscript{\rm 1,2}, Sherif Abdelkarim\textsuperscript{\rm 1},
  Deyao Zhu\textsuperscript{\rm 1},  Mohamed Elhoseiny\textsuperscript{\rm 1}\\
  \textsuperscript{\rm 1}{King Abdullah University of Science and Technology} \\ 
\textsuperscript{\rm 2}{ Indian Institute of Technology} \\
\texttt{\{jun.chen,deyao.zhu,mohamed.elhoseiny\}@kaust.edu.sa} \\
 \texttt{aagarwal@ma.iitr.ac.in}, \texttt{sherif.abdelkarim91@gmail.com} 
}

\twocolumn[{
\renewcommand\twocolumn[1][]{#1}%

\maketitle
\begin{center}
    \centering
    \includegraphics[width=0.85\linewidth]{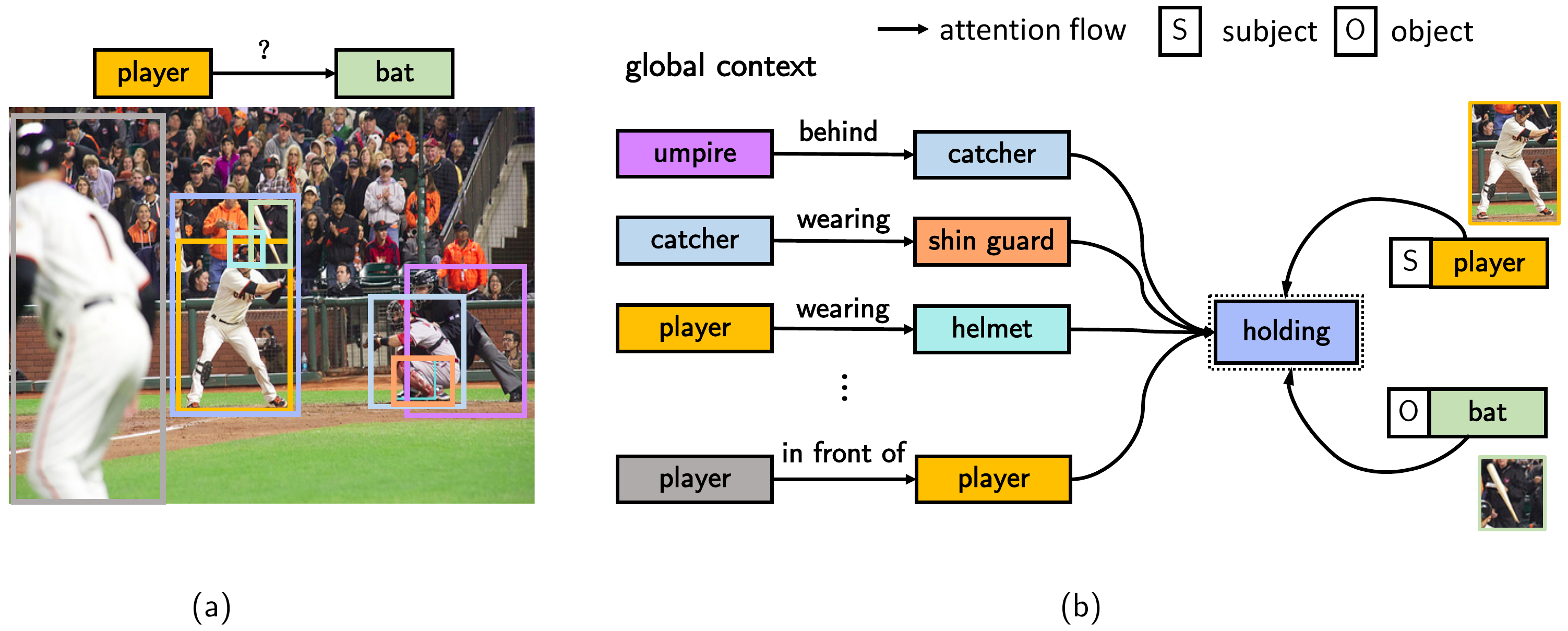}
    \captionof{figure}{(a) Given an image with several annotated objects, the objective is to predict the visual relationship between the image region of ``player" and ``bat". (b) It illustrates our message-passing strategy: we attentively aggregate visual features from all the triplets in the global context, the subject, and object to the ``holding" feature via attention. All the boxes above denote the corresponding visual features.}
    \label{fig:teaser}
\end{center}
}]

\begin{abstract}

The visual relationship recognition (VRR) task aims at understanding the pairwise visual relationships between interacting objects in an image. These relationships typically have a long-tail distribution due to their compositional nature. This problem gets more severe when the vocabulary becomes large, rendering this task very challenging. This paper shows that modeling an effective message-passing flow through an attention mechanism can be critical to tackling the compositionality and long-tail challenges in VRR. The method, called RelTransformer, represents each image as a fully-connected scene graph and restructures the whole scene into the relation-triplet and global-scene contexts. It directly passes the message from each element in the relation-triplet and global-scene contexts to the target relation via self-attention. We also design a learnable memory to augment the long-tail relation representation learning. Through extensive experiments, we find that our model generalizes well on many VRR benchmarks. Our model outperforms the best-performing models on two large-scale long-tail VRR benchmarks, VG8K-LT (+2.0\% overall acc) and GQA-LT (+26.0\% overall acc), both having a highly skewed distribution towards the tail. It also achieves strong results on the VG200 relation detection task. Our code is available at \url{https://github.com/Vision-CAIR/RelTransformer}.

\end{abstract}

\section{Introduction}
\label{sec_intro}



The Visual Relationship Recognition (VRR) task goes beyond recognizing individual objects by comprehensively understanding relationships between interacting objects in a visual scene. Owing to the enriched scene understanding provided by VRR, it benefits various other vision tasks such as image captioning (e.g., \cite{visualRelationImageCaption, sceneGraphImageCaption,elhoseiny2017sherlock}), VQA (e.g., \cite{graphVQA,gqa}), image generation (e.g.,  \cite{imageGenerationFromScene}),  and 3D scene synthesis (e.g., \cite{3dscene}). However, due to the imbalanced class distribution in many VRR datasets \cite{visualGenome, gqa}, predictions of the most existing models are dominated by the head/frequent relations, lacking generalization on tail/low-shot relationships.


Many previous approaches characterize the VRR problem under a graph scenario. Popular graph-based methods iteratively pass massages from other direct or indirect nodes to the relation along with the structure of the graph using the long short-term memory \cite{motifNet,messagePassing,vctree}, or graph attention networks \cite{graph-rcnn,gpsNet}. However, the graph structure may implicitly constraint the relation to focus on its nearby neighbors. This phenomenon has been observed in recent works \cite{alon2020bottleneck,wu2020comprehensive}, showing that graph neural networks incline to pay most attention to the local surrounding nodes but could not benefit much from distant nodes~\cite{alon2020bottleneck}, and node representations will become indistinguishable if there are many layers \cite{wu2020comprehensive}. Such problems can also be seen in long short-term memory networks due to their iterative message-passing learning nature. However, the target relation can also benefit a lot from the distant nodes. e.g., when we predict the ``holding" relationship between ``player" and ``bat" in Fig. \ref{fig:teaser}, the distant objects such as ``catcher" and ``umpire" can provide a context that the ``player" is in a baseball game, and those can help the model better predict ``holding" relationship.


To alleviate the aforementioned problems, we propose to adapt self-attention mechanisms originally introduced in Transformer \cite{vaswani2017attention} to tackle the VRR challenges.
Self-attention can be viewed as a non-local mean operation \cite{buades2005non}, which computes the weighted average of all the inputs. When it applies to the VRR problem, it assumes that the relation has a full connection with all the other nodes in the graph and directly passes the messages among them via attention. In contrast to GNN/LSTM approaches, this strategy can allow the relation to have a larger attention scope and pass the message regardless of the graph structure or spatial constraints. It also avoids the valuable information from distant nodes being suppressed by nearby neighbors. Hence, each relation can selectively attend to its relevant features without spatial constraints and learn a richer-contextualized representation, which can benefit the long-tail visual relationship understanding.


In our approach, dubbed as RelTransformer, we reconstruct the scene graph into the relation-triplet and global-scene context as we demonstrated in the Fig. \ref{fig:teaser}. 
The relation triplet here refers to the target relation and its referred subject and object, such as $\langle \text{player}, \text{holding}, \text{bat}\rangle$ in the figure. 
The global context represents all the relation triplets that are gathered for each appearing relation. We directly connect the target relation ``holding" with every element from the relation triplet and global context, and pass their information to the target relation via self-attention.
Furthermore, since the long-tail relations tend to be amenable to forgetting, we also propose a novel memory attention module to augment the relation representation with external persistent memory vectors, as we will detail later.



We showcase the effectiveness of our model on VG200 \cite{visualGenome}  and two recently proposed large-scale long-tail VRR benchmarks, GQA-LT \cite{ltvrd} and VG8K-LT \cite{ltvrd}. GQA-LT and VG8K-LT scale the number of relation types up to 300 and 2,000 compared to only 50 relation types in VG200. These two benchmarks are highly skewed (e.g., the VG8K-LT benchmark ranges from 14 to 618,687 examples per relation type) and offer us a suitable platform for studying long-tail VRR problems. Our approach achieves the state-of-the-art on those three datasets in our experimental results, demonstrating its effectiveness. We also conducted several ablative experiments and showed the usefulness of each component design in RelTransformer.


\section{Background}

\begin{figure*}[t!]
\centering
\includegraphics[width=0.76\linewidth]
                  {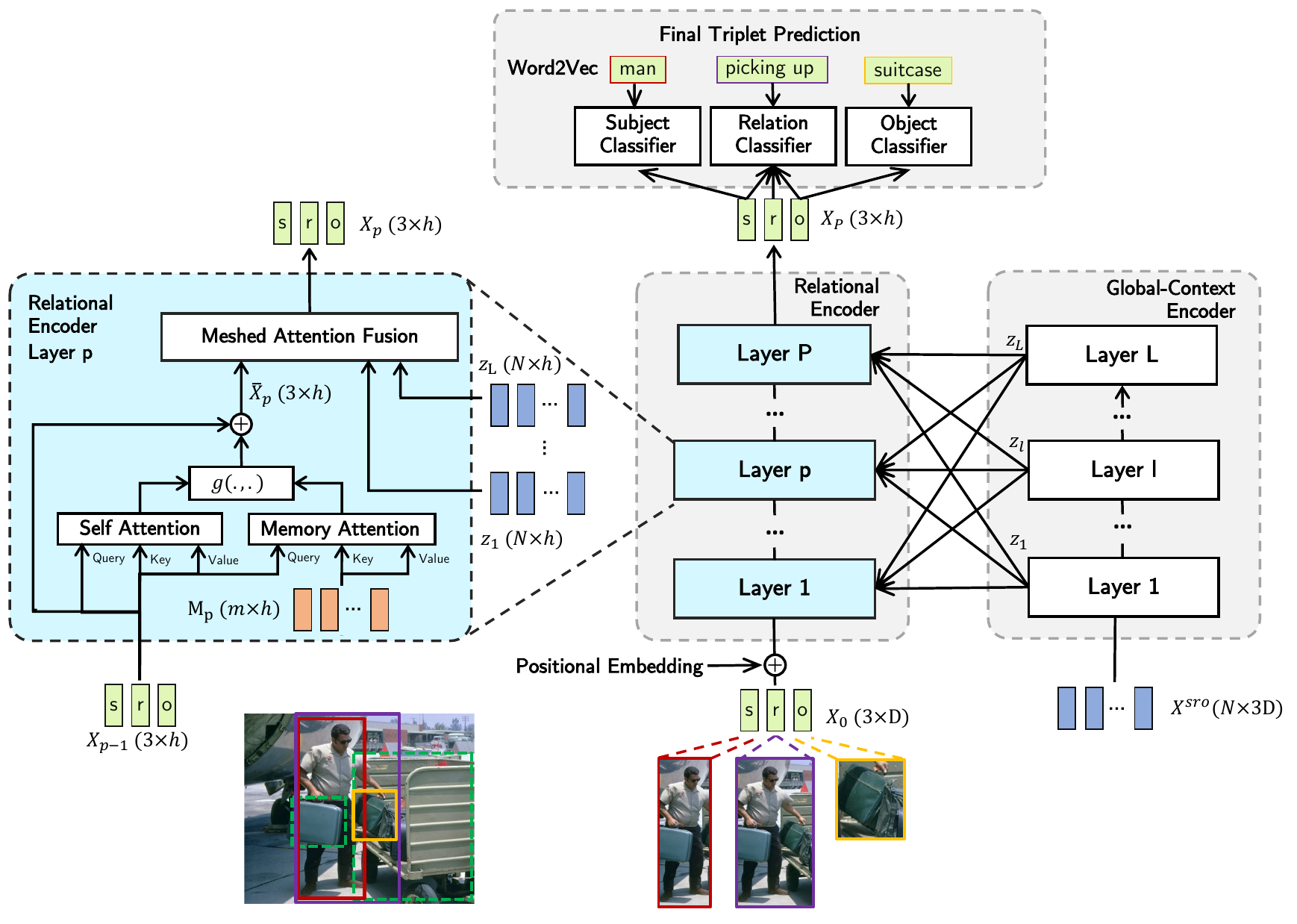}
\caption{The architecture of our RelTransformer. It comprises a multi-layer global-context and a relational encoder. We display the detailed operations from a relational encoding layer in the left, which include memory attention and meshed attention fusion modules.}
\label{fig:main_model}
\end{figure*}

\noindent \textbf{Visual Relationship Recognition.} Correct visual relation prediction requires having a comprehensive understanding of the image contents, which guides many successful works in the literature. Early works employ RNN models to construct a global context by aggregating the node and edge features via iterative message passing such as \cite{messagePassing,motifNet,vctree,li2017scene,permutationInvariant}. e.g.,  VCTree \cite{vctree} composes a dynamic tree structure to organize the object orders and apply TreeLSTM \cite{tai2015improved} to aggregate features. 
There are also several graph convolutional network (GCN) \cite{gcn} approaches \cite{graph-rcnn,attentiveRelNet,chen2019knowledge,subgraphSceneGeneration}, which attempt to learn different importance weights to the neighborhood nodes. Lin \textit{et al.} \cite{gpsNet} extend the graph attention to also capture the node-specific contextual information and encode the edge direction information. More recently, there is also emerging a Transformer-based approach \cite{rtn}, which models the pairwise interaction among nodes and edges in two separate Transformer networks. Our model differs from it mainly in two aspects: a) we have a different message-passing flow in which we specifically aggregate relation features from the relation-triplet and global-context information. b) We further design an effective memory attention module for augmenting the long-tail relation representation.

\noindent \textbf{Long-Tail Visual Relationship Recognition.} Long-tail problem is very severe in visual relation recognition (VRR) \cite{lsvru,ltvrd,gpsNet}. There are mainly two approaching directions to alleviate this problem. The first one is semantically guided visual recognition. In this case, language models \cite{word2vec} are employed for zero-shot or few-shot recognition \cite{frome2013devise,norouzi2014zero}. There are also several VRR works \cite{lsvru,ltvrd,graph-rcnn,yu2017visual} using the language priors as a guidance to learn relation features, which can derive a better classification on long-tail classes. The second direction is to apply the strategies that have been designed for unbalanced object detection, including various class imbalance loss functions (e.g. weighted cross entropy, focal loss \cite{focalLoss}, equalization loss \cite{eql}), sampling strategies (e.g., under-sampling \cite{underSampling},  over-sampling \cite{overSampling} and class-balanced sampling \cite{balanceSampling}), data augmentation \cite{cutmix,ltvrd}, meta learning \cite{ha2016hypernetworks, wang2017learning}, counterfactual learning \cite{tang2020unbiased}, memory modules \cite{oltr} and decoupling methods \cite{decoupling}. In our long-tail VRR experiment, we also leverage the language models and many aforementioned long-tail strategies to better classify long-tail relation types. But we focus on understanding their collaborative effect with our message-passing mechanism and how they influence head and tail predictions.  

\section{Approach}



\subsection{Problem Definition}
An image can be decomposed into a scene graph $G = (N, E)$, where each node $(n_i \in N)$ represents an object and each edge $(e_i  \in E)$ represents the spatial or semantic relationships between two interacting objects. We denote the visual relationship between a subject $n_s$ and an object $n_o$ as $r$. In the visual relationship recognition (VRR) task, the goal is to predict $r$ between the given $n_s$ and $n_o$.

\begin{equation}
y^{r} = f(b^s, b^o, b^r, I) 
\label{problem_define}
\end{equation}
where $b^s$, $b^o$, and $b^r$ are the subject, object and relationship bounding boxes. $b^r$ is obtained by the minimum enclosing region of $b^s$ and $b^o$. $y^{r}$ is the relationship label. $I$ denotes the arbitrary information such as image raw RGB pixel features. $f$ is the inference model.

\subsection{RelTransformer Architecture}

RelTransformer mainly comprises two components, an $L$-layer global-context encoder and a $P$-layer relational encoder. The overall architecture can be seen in Fig. \ref{fig:main_model}. Given an image, we first extract the object and relation features via a faster R-CNN detection network \cite{ren2015faster}. Those features are grouped together as a tuple of triplets 
$(\langle s_1 , r_1, o_1\rangle,\dots,\langle s_N, r_N, o_N\rangle)$
according to their spatial or semantic relationships provided in the dataset, where each $s_i$, $r_i$ and $o_i$ $\in$ $\mathbb{R}^{1\times D}$ and  $N$ is the number of triplets. We first feed all the triplets into a multi-layer global-context encoder to learn a scene-contextualized representation.
Then we concentrate on learning the target relation $r_i$ representation from the context of $\langle s_i , r_i, o_i \rangle$ in the relational encoder. An external memory module is also introduced here to augment the long-tail relation representations. Finally, we employ a meshed connection (see Fig. \ref{fig:main_model}) to integrate global-context features into the relation representation learning.

\noindent \textbf{Global-Context Encoder.} We employ an $L$-layer Transformer \cite{vaswani2017attention} to model the global-context information and learn their pairwise relationships. 
We first concatenate each triplet from $(\langle s_1 , r_1, o_1\rangle,\dots,\langle s_N, r_N, o_N\rangle)$ together into a compact representation as $X^\text{sro} = (x^\text{sro}_1 \dots x^\text{sro}_{N}$) where  $X^\text{sro} \in \mathbb{R}^{N \times 3D}$. We then feed $X^\text{sro}$ into the global-context encoder in a permutation-invariant order.




The Transformer \cite{vaswani2017attention} is a stack of multi-headed self-attention (MSA) and MLP layers. Its core component is the self-attention as defined in Eq. \ref{eq:self-att}. In each Transformer encoding layer, the multi-head self-attention repeats the self-attention multiple times and concatenate the results together; the result is then projected back to the same dimensionality. After that, the result is fed into an MLP network and produces each layer's output. 

\begin{equation} 
f_\text{sa}(Q, K, V) =\text{softmax}\left( \frac{(W_{q}Q) (W_{k}K)^{\top}}{\sqrt{d}}\right) W_{v}V
\label{eq:self-att}
\end{equation}

\noindent where $Q$, $K$ and $V \in \mathbb{R}^{t \times h}$ are query, key and value vectors. $t$ is the number of input tokens and $h$ is the hidden size. $d$ is a scaling factor. $W_q$, $W_k$ and $W_v$ are the learnable weight parameters. 

We feed $X^\text{sro}$ into the global-context encoder, each layer $l$ outputs a contextualized representation $z_l$. We gather them together as $Z$ = ($z_{1}, \dots, z_{L}$) where $z_{l} \in \mathbb{R}^{N \times h}$.


\noindent \textbf{Relational Encoder.} It has multiple functions: 1) it specifically aggregates the relation representation from its referred subject and object via self-attention 2) it incorporates a persistent memory to augment the relations with ``out-of-context" information, which could especially benefit the long-tail relations. 3) it also has a ``bridging" role to aggregate each global-context encoding layer's output into the relation representation through a meshed attention module. The detailed operating procedure is described as follows.



We first add a learnable positional embedding \cite{bert} to each position of $s_i$, $r_i$, and $o_i$ in order to distinguish their semantic differences in the input sequence, and we denote this sequence as $X_{p-1}$ in the layer $p$. We apply the self-attention defined in Eq. \ref{eq:self-att} on $X_{p-1}$ in each layer $p$ and model their pairwise relationship. Their result will be $X^{att}_p$ as shown in Eq. \ref{sf1}. 
\begin{equation}
        X^{\text{att}}_p = f_\text{sa}(X_{p-1},  X_{p-1}, X_{p-1})
        \label{sf1}
\end{equation}

\noindent \textbf{Memory Augmentation.} The trained model can easily forget long-tail relations because the model training is dominated by the instance-rich (or head) relations, and hence it tends to underperform on lower-frequent relations. Also, the self-attention is limited to attending to the features only from the tokens in an input sequence; hence, each relation only learns a representation based on the immediate context. To alleviate this issue, we propose a novel memory attention module motivated by several successful persistent memory ideas \cite{persistentMemory,zhao2020speech} in the literature. We denote a group of persistent and differentiable memory vectors as $M$. Each time the relation passes its features to $M$ and retrieves the information from $M$ via attention. The memory here captures the information not dependent on the immediate context; instead, it is shared across the whole dataset \cite{persistentMemory}. Through this way, long-tail relations are able to access the information (e.g., from other relations or itself in different training steps) with relevance to itself, and they can augment the target relation with the useful ``out-of-context" information in a well-trained model.




To compute this memory, we first randomly initialize $m$ memory vectors in each relational encoding layer $p$ as $M_p \in \mathbb{R}^{m \times h}$. We then compute the memory attention between the input feature with $M_p$, we treat $X_{p-1}$ as the query, and $M$ here is the Key and Value. Same self-attention operation is applied here in Eq. \ref{sf2} and we can obtain $X^{\text{mem}}_p$. The memory is directly updated via SGD. 
\begin{equation}
        X^\text{mem}_p = f_\text{sa}( X_{p-1},M_{p-1} ,M_{p-1})
        \label{sf2}
\end{equation}

To aggregate $X^{\text{mem}}_p$ into $X_p^\text{att}$, we design a fusion function as $g(x,y)$ in Eq. \ref{fusion}. $g(x,y)$ is a attention gate and determines how to effectively combine two input features. It computes the complementary attention weights to each input and weighted combine them as the output. Through this fusion function in combination with a skip connection, we can get the fused feature as $\bar{X}_p$ = $g(X_p^\text{att}, X_p^{\text{mem}})$ + $X_{p-1}$.
\begin{equation}
\begin{split}
   g(x,y) &=  \alpha \odot x  + (J-\alpha) \odot y\\
   \alpha &= \sigma( W[x ; y] + b)
\end{split}
\label{fusion}
\end{equation}
 where $W$ is a 2D $\times$ D matrix. $b$ is a bias term. $[;]$ denotes the concatenation. $\odot$ is the Hadamard product. $J$ is an all-one matrix with the same dimensions as $\alpha$. 

\noindent \textbf{Meshed Attention Fusion.} The features from different global-context encoding layers capture different vision granularity, and leveraging features from all of them has shown to be better than the one only from the last encoding layer~\cite{m2transformer,visualgpt}. Therefore, we adopt a meshed connection in our model and contribute each layer's output $z_l$ to the relation representation. To compute the meshed attention, we first compute the cross-attention between $\bar{X}_p$ and each global-context encoding output in ($z_1,\dots,z_L$); Its attention output is fused with $\bar{X}_p$ through Eq. \ref{fusion}. Their results are averaged up for each layer. We then project the averaged output in an MLP network and incorporate a skip connection to compute the final fused relation representation $X_p$ in Eq. \ref{crossfusion} as the layer output.


\begin{equation}
\begin{split}
     z_p^l  =& f_\text{sa}(\bar{X}_p, z_l, z_l)\\
   {X}_p =&  \text{MLP}(\frac{1}{L} \sum_{l=1}^L g(\bar X_p, z_p^l))+\bar{X}_p
\end{split}
\label{crossfusion}
\end{equation}


\begin{table*}[t!]
\centering
\scalebox{0.86}{
\begin{tabular}{ ll| cccc|cccc}
 \multicolumn{2}{l}{} & \multicolumn{4}{c}{VG8K-LT} & \multicolumn{4}{c}{GQA-LT} \\

 \multirow{2}{*}{Architecture} & \multirow{2}{*}{Learning Methods} & many &  medium &  few &  all &  many &  medium &  few &  all \\
  &  & 100 &  300 &  1,600 & 2,000&  16 &  46 &  248 &  310 \\
\toprule

LSVRU & VilHub \cite{ltvrd} & 27.5 & 17.4 & 14.6 & 15.7 & \underline{$\text{63.6}$} & 17.6 & 7.2 & 11.7\\
LSVRU & VilHub + RelMix \cite{ltvrd}  & 24.5 &16.5 &14.4& 15.4 & 63.4 &14.9& 8.0 &11.9 \\

LSVRU & OLTR \cite{oltr} & 22.5 & 15.6 & 12.6 & 13.6 & 63.5 & 15.0 & 8.2 & 12.1 \\ 
LSVRU & EQL \cite{eql} & 22.6 & 15.6 & 12.6 & 13.6 & 62.3 & 15.8 & 6.6 & 10.8 \\

LSVRU & $\text{Counterfactual }^{\sharp}$ \cite{tang2020unbiased} & 12.1 & 25.6 & 14.9 & 17.1 & 38.6 & 38.0 & 9.4 & 15.2\\


\midrule
 LSVRU& CE &22.2 & 15.5 & 12.6 & 13.5 & 62.6 & 15.5 & 6.8 & 11.0\\
\rowcolor{Gray2}
 RelTransformer (ours)& CE & \textbf{$\text{26.8}$} &  \textbf{$\text{18.6}$}   & \textbf{15.0} & \textbf{16.1} & \textbf{63.4} & \textbf{16.6} & \textbf{7.0} & \textbf{11.2} \\
\midrule

LSVRU & Focal Loss \cite{focalLoss} & 24.5 & 16.2 & 13.7 & 14.7 & 60.4 & 15.7 & 7.7 & 11.6 \\
\rowcolor{Gray2}
RelTransformer (ours) & Focal Loss & \textbf{30.5} & \textbf{22.8} & \textbf{14.8} & \textbf{16.8} & \textbf{61.9} & \textbf{16.8} & \textbf{8.3} & \textbf{12.2}\\
\midrule
LSVRU & DCPL \cite{decoupling}& 34.3 & 15.4 & 12.9 & 14.4& \textbf{61.4} & 23.6 & 7.6 &12.7 \\
\rowcolor{Gray2}
RelTransformer (ours) & DCPL  & \textbf{\underline{37.3}} & \underline{\textbf{27.6}}& \underline{\textbf{16.5}} & \underline{\textbf{19.2}}& 58.4& \textbf{38.6}& \textbf{13.2}&  \textbf{19.3} \\
\midrule
LSVRU  &  WCE &  35.5 & 24.7 & 15.2 & 17.2 & 53.4 & 35.1 & 15.7 & 20.5 \\ 
\rowcolor{Gray2}
RelTransformer (ours) & WCE &  \textbf{36.6}  & \textbf{$\text{27.4}$}& \textbf{$\text{16.3}$} & \textbf{$\text{19.0}$}&    \underline{$\textbf{63.6}$} & \underline{\textbf{$\text{59.1}$}} & \underline{\textbf{$\text{43.1}$}} &\underline{\textbf{$\text{46.5}$}}

\end{tabular}
}
\newline
\caption{Average per-class accuracy in relation prediction on VG8K-LT and GQA-LT datasets. We evaluate the average per-class accuracy for many, medium, few, and all classes. The best performance for each column is \underline{underlined}. $\sharp$ denotes our reproduction. Learning methods include various class-imbalance loss functions, augmentation and counterfactual approaches. Our model is marked in \colorbox{Gray2}{gray}.}
\label{main_results}
\end{table*}

\noindent \textbf{Final Triplet Prediction.} In the last layer $P$ of the relational encoder, we extract subject $x_s$, relation $x_r$ and object $x_o$ from $X_P$ accordingly. In the prediction stage, we leverage the language prior knowledge following previous work \cite{lsvru}, and represent each ground-truth label in their Word2Vec \cite{word2vec} embeddings, we project them into hidden representations with a 2-layer MLP in the classifier. Finally, we maximize their cosine similarity with $x_s$, $x_r$, and $x_o$ individually during the training.

\section{Experiments}

\subsection{Datasets}

We evaluate our model on the VG200 dataset and two large-scale long-tail VRR datasets, named GQA-LT \cite{ltvrd} and VG8K-LT \cite{ltvrd}.

\noindent \textbf{GQA-LT.} This dataset contains $72,580$ training, $2,573$ validation and $7,722$ testing images. Overall, it contains $1,703$ objects and $310$ relationships. The GQA-LT has a heavy ``long-tail" distribution where the example numbers per each class range from merely $1$ to $1,692,068$. 

\noindent \textbf{VG8K-LT.} It is collected from Visual Genome (v1.4) \cite{visualGenome} dataset, containing $97,623$ training, $1,999$ validation and $4,860$ testing images. It covers $5,330$ objects and $2,000$ different relation types in total, in which least frequent objects/relationships only have $14$ examples and the most frequent ones have $618,687$ examples. 

\noindent \textbf{VG200.} This dataset has been widely studied in the literature \cite{messagePassing, lsvru, zhang2019graphical}. It contains 50 relationships, and the category frequency in this dataset is considerably more balanced than that in GQA-LT and VG8K-LT. We follow the same data split as in \cite{lsvru} in our experiment.

\subsection{Experimental Settings}

\noindent\textbf{GQA-LT \& VG8K-LT Baselines.} We compare RelTransformer with several state-of-the-art models. The most popular models in this benchmark are implemented based on LSVRU \cite{lsvru} framework. To improve over the long-tail performance, baseline models usually are combined with the following strategies: 1) class-imbalance loss functions such as weighted cross entropy (WCE), equalization loss (EQL) \cite{eql}, focal loss \cite{focalLoss}, and ViLHub loss \cite{ltvrd}. 2) relation augmentation strategies such as RelMix \cite{ltvrd} to augment more examples in long-tail relations. 3) Decoupling \cite{decoupling} which decouples the learning procedure into representation learning and classification. 4) Counterfactual \cite{tang2020unbiased}, which alleviates the biased scene graph generation via the counterfactual learning. 5) OLTR \cite{oltr}, which has memory module with augmented attention.

\noindent\textbf{VG200 Baselines.} We compare with several strong baselines including Visual Relationship Detection \cite{vrd}, Message Passing \cite{messagePassing}, Associative Embedding \cite{associative}, MotifNet \cite{motifNet}, Permutation Invariant Predication \cite{permutationInvariant}, LSVRU \cite{lsvru}, relationship detection with graph contrastive loss (RelDN) \cite{zhang2019graphical},  GPS-Net \cite{gpsNet}, Visual Relationship Detection with Visual-Linguistic Knowledge (RVL-BERT) \cite{chiou2021visual} and Relational Transformer Network (RTN) \cite{rtn}.

\begin{table}[t]
    \centering
    \scalebox{0.9}{
    \begin{tabular}{l ccc }
    \multirow{2}{*}{Models} & \multicolumn{3}{c}{PRDCLS}\\
    & R@20 & R@50 & R@100\\
    \toprule
     VRD \cite{vrd}& - & 27.9 & 35.0  \\
     Message Passing \cite{messagePassing}& 52.7 & 59.3 & 61.3  \\
     Associative Embedding \cite{associative} & 47.9 & 54.1 & 55.4\\
     MotifNet (Left to Right) \cite{motifNet} & 58.5 & 65.2 & 67.1  \\
     Permutation Invariant \cite{permutationInvariant} &  - & 65.1 & 66.9 \\
     LSVRU \cite{lsvru}  & 66.8 & 68.4 & 68.4 \\
     RelDN \cite{zhang2019graphical}  & 66.9 & 68.4 & 68.4\\
     Graph-RCNN \cite{graph-rcnn}  & - & 54.2 & 59.1 \\
     VCTREE-SL  & 59.8 & 66.2 & 67.9 \\
     GPS-Net \cite{gpsNet} & 60.7 & 66.9 & 68.8  \\
     RVL-BERT \cite{chiou2021visual} &-  &62.9& 66.6\\
     RTN \cite{rtn} & 68.3& 68.7 &68.7\\
     \midrule
     RelTransformer (ours) & \textbf{68.5} & \textbf{69.7} & \textbf{69.7} \\
    \end{tabular}
    }
    \caption{Relation prediction on VG200 dataset. 
    }
    \label{vg200_result}
\end{table}

\noindent\textbf{Evaluation Metrics.} For the GQA-LT and VG8K-LT datasets, we report the average per-class accuracy, which is commonly used for long-tail evaluation \cite{eql, decoupling, ltvrd}. Following the same evaluation setting as \cite{ltvrd}, we split the relationship classes into the many, medium, and few based on the relation frequency in the training dataset as shown in Table \ref{main_results}. For the VG200 dataset, following the previous evaluation setting in \cite{lsvru,Suhail_2021_CVPR}, we measure the Recall@k and mean Recall@K for predicate classification (PRDCLS), which is to predict the relation labels given the ground truth boxes and labels of the subject and object.


\subsection{Quantitative Results}

\noindent\textbf{GQA-LT and VG8K-LT Evaluation.} We present our results for GQA-LT and VG8K-LT datasets in Table \ref{main_results}. There is a clear performance improvement over all the baselines with the addition of RelTransformer, especially on the med and few classes. The combination of RelTransformer with WCE improves the med and few category in GQA-LT by a considerable margin of $\approx$20\% compared to all the baselines. This huge gain can be attributed to the weighted assignment of different classes when WCE loss is applied. This further refines the attention weights assigned to different classes in the global context and hence helps the overall performance. While previous works \cite{tang2020unbiased} improved the tail performance at the cost of head class accuracy, RelTransformer consistently improves on the tail as well as the head as seen from the table, underlying the effectiveness of our model.

\begin{figure*}[t]
\begin{minipage}{0.5\textwidth}
\centering
 \includegraphics[width=1.0\linewidth]
    {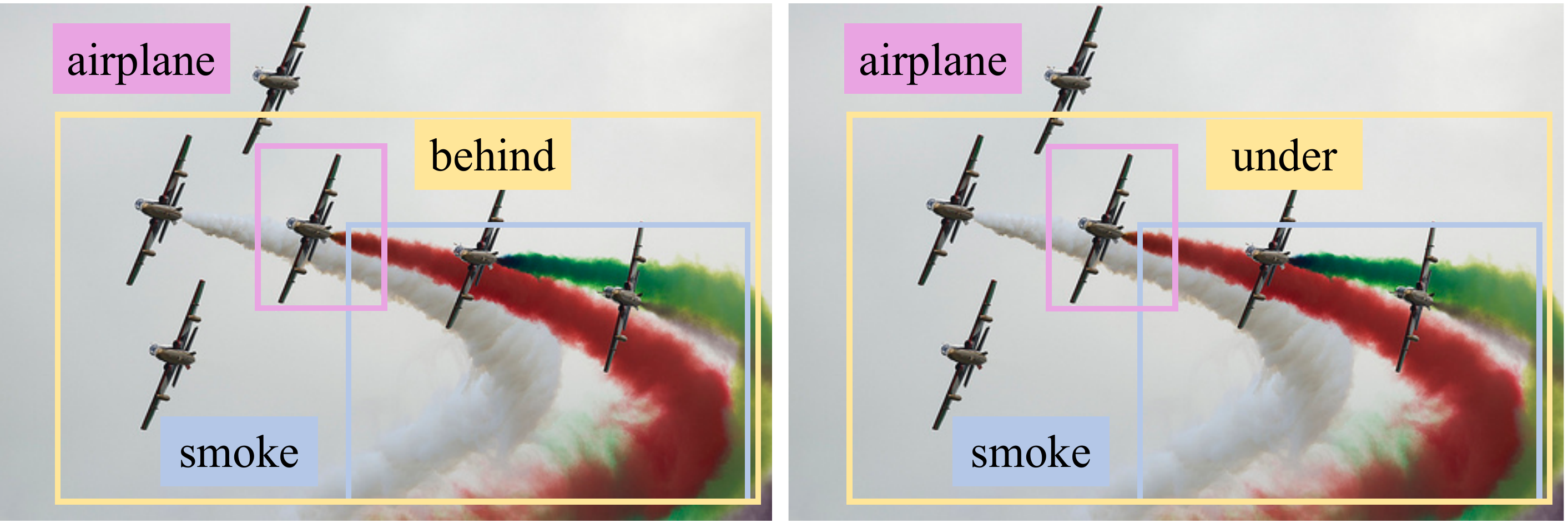}   
\end{minipage} \hfill
\begin{minipage}{0.5\textwidth}
\centering
 \includegraphics[width=1.0\linewidth]
    {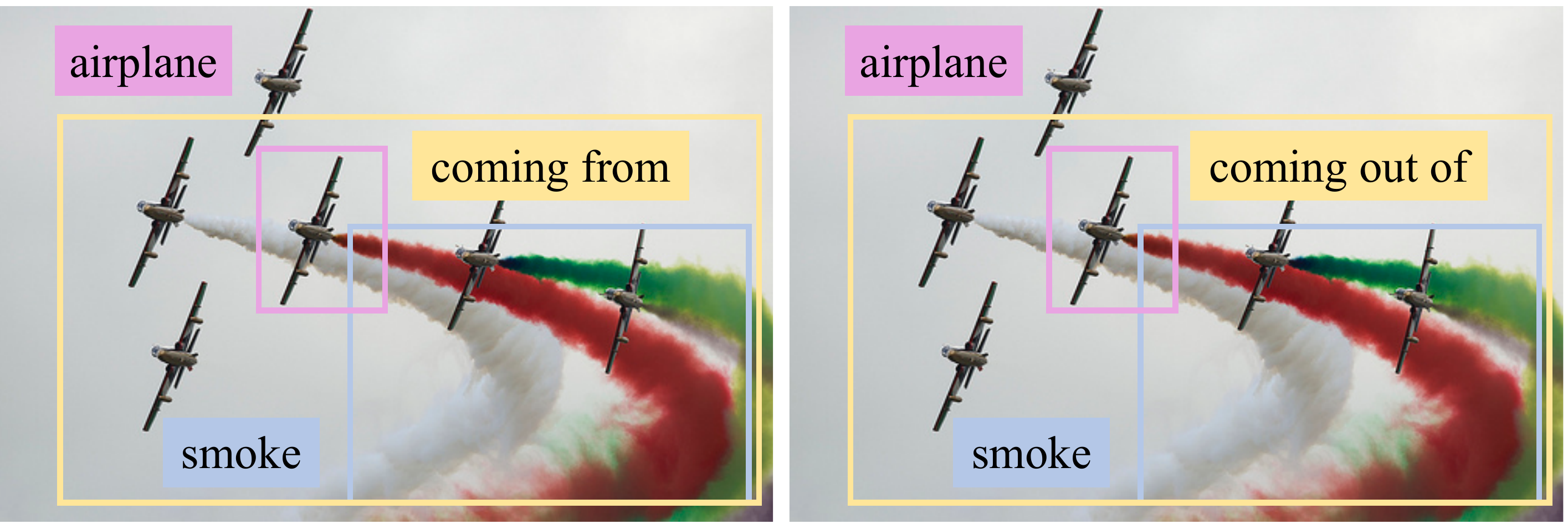}   
\end{minipage} \hfill
\begin{minipage}{0.5\textwidth}
\centering
 \includegraphics[width=1.0\linewidth]
    {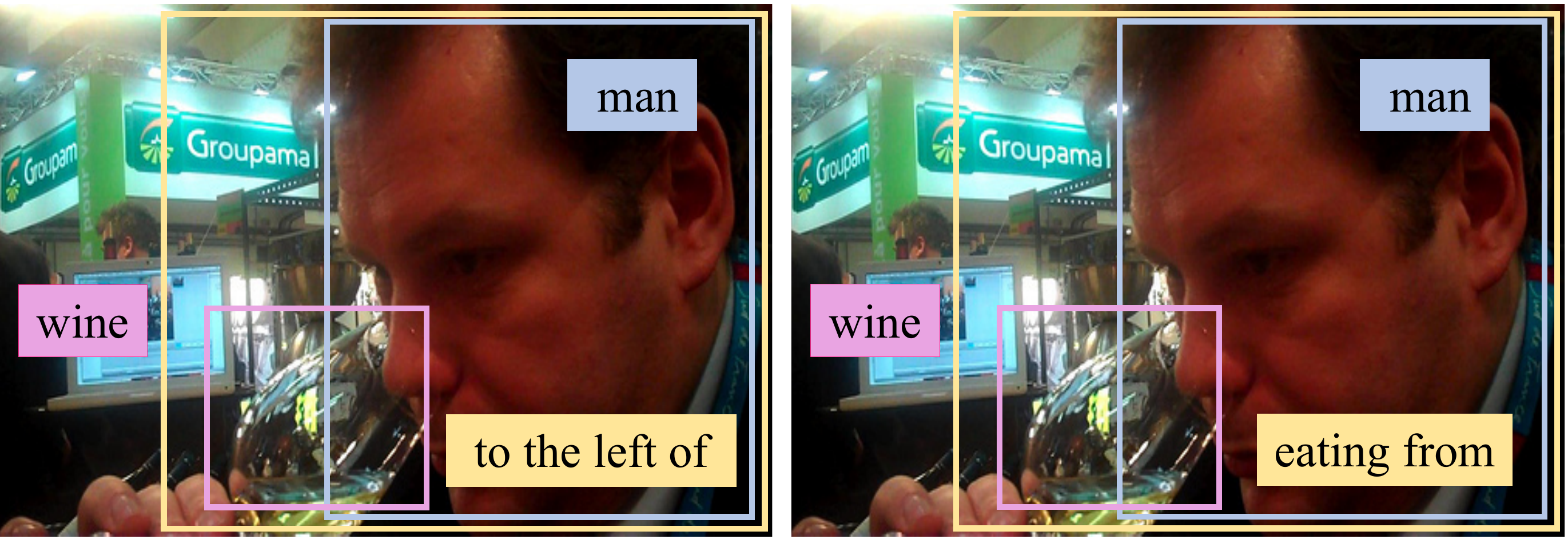} 
    \title{ \hspace{20pt}  LSVRU  \hspace{70pt}  RelTransformer }
\end{minipage} \hfill
\begin{minipage}{0.5\textwidth}
\centering
 \includegraphics[width=1.0\linewidth]
    {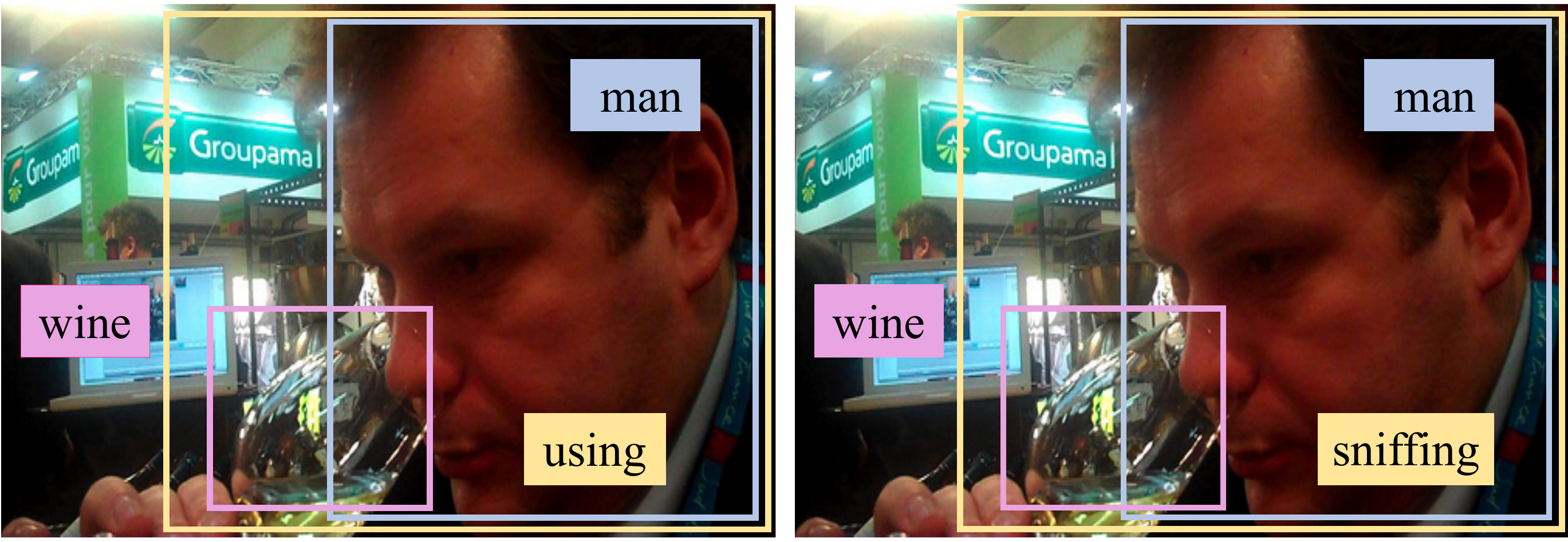}   
    \title{  LSVRU (WCE)  \hspace{50pt}      RelTransformer (WCE) }
\end{minipage} \hfill
\caption{Long-tail relationship recognition visualization. We compare our model with LSVRU in the combination of CE and WCE loss functions. Yellow color denotes the relation, blue denotes the subject, and pink denotes the object.}
\label{demonstrations}
\end{figure*}

For VG8K-LT, we also see performance gains with the addition of RelTransformer on all baselines across ``many", ``medium" and ``few" categories. A considerable improvement of $\approx$5\% can be seen when RelTransformer is combined with DCPL \cite{decoupling}, performing the best for the VG8K-LT dataset. While we see consistent improvements with the addition of RelTransformer for the VG8K-LT dataset, the improvement margins are certainly lower as compared to GQA-LT as seen in Table \ref{main_results}. This is due to the more challenging nature of VG8K-LT, containing 2000 relation classes compared to 300 classes present in GQA-LT. Some qualitative examples with the addition of RelTransformer can be seen in Fig. \ref{demonstrations}.

\noindent\textbf{VG200 Evaluation.} We also evaluate our model on VG200 dataset in Table \ref{vg200_result}. We compare with many different message-passing approaches in our baselines including RNN-based \cite{vctree,motifNet}, GCN-based \cite{graph-rcnn,gpsNet} and Transformer-based \cite{rtn,chiou2021visual}. These two Transformer-based approaches either only focus on the relational triplet context \cite{chiou2021visual} or neglect it completely \cite{rtn}. Our model differs from them with a different message-passing strategy, different context construction and a novel memory attention. The experimental results show that our method can better exploit the relational features, and we improve over the best-performing baselines by $0.2\%$ on R@20, $1.0\%$ on R@50, and $0.9\%$ on R@100.

\noindent \textbf{Mean Recall@K on VG200.} We evaluate the mean recall@k performance on the VG200 dataset for elation predication, and compared RelTransformer with several strong baselines such as VCTREE and Motif with both cross-entropy and EBM losses \cite{Suhail_2021_CVPR}. The results are summarized in Table~\ref{vg200_mreall}. We observe that RelTransformer can outperform all the baselines on mR@(20, 50, 100) while only being combined with cross-entropy loss, which shows its robustness on other data-imbalanced datasets.



\begin{table}[t]
    \centering
    \scalebox{0.9}{
    \begin{tabular}{l cccc }

    \multirow{2}{*}{Models} &\multirow{2}{*}{Method} &  \multicolumn{3}{c}{PRDCLS}\\
    & & mR@20 & mR@50 & mR@100\\
    \toprule
     IMP  & CE & 8.85 & 10.97 & 11.77 \\
     IMP  & EBM \cite{Suhail_2021_CVPR}  & 9.43 & 11.83 & 12.77 \\
     Motif & CE &12.45 & 15.71 & 16.8 \\
     Motif & EBM &14.2 & 18.2 & 19.7 \\
     VCTREE  & CE & 13.07 & 16.53 & 17.77 \\
     VCTREE  & EBM & 14.17 & 18.02 & 19.53 \\
     \midrule
    Ours & CE & \textbf{18.51} & \textbf{19.58} & \textbf{20.19}\\
    \end{tabular}}
    \caption{Mean Recall@K Performance on VG200 Dataset.}
    \label{vg200_mreall}
\end{table}

\begin{table*}[t]
\centering
\scalebox{0.9}{
\begin{tabular}{ cc| ccccccccc}
 
 \multicolumn{2}{c}{} & \multicolumn{3}{c}{many} & \multicolumn{3}{c}{medium} & \multicolumn{3}{c}{few} \\
Architecture & Learning Methods & SO &  SR &  OR &   SO &  SR & OR   &  SO & SR & OR \\
\toprule
LSVRU & VilHub & 40.5 & 32.8 & 33.7  & 25.7 & 14.2 & 13.9  & 10.2 & 5.3 & 5.2 \\
\midrule
LSVRU & CE &38.6 & 30.3 & 31.5  & 21.8 & 11.3 & 10.8  & 7.5 & 4.3 & 4.2 \\
RelTransformer & CE & \underline{\textbf{$\text{54.2}$}} & \underline{\textbf{$\text{46.6}$}} &  \underline{\textbf{$\text{47.2}$}} & \underline{\textbf{$\text{37.4}$}} &  \underline{\textbf{$\text{20.8}$}} & \underline{\textbf{$\text{21.8}$}} & \underline{\textbf{$\text{16.1}$}} & \textbf{8.6}  & \textbf{7.7}\\
\midrule
LSVRU & Focal Loss & 39.2 & 31.1 & 32.3 & 23.2 & 11.9 & 11.5 & 8.2 & 4.3  & 4.2\\
RelTransformer & Focal Loss & \textbf{49.9} & \textbf{41.7}& \textbf{42.5} & \textbf{32.2} & \textbf{17.7} & \textbf{8.0} & \textbf{13.1} & \textbf{7.0}  & \textbf{6.4}\\
\midrule
LSVRU  &  WCE &  18.3 & 17.3 & 17.2  & 13.7 & 9.4 & 9.4  & 7.1 & 4.2 & 3.6  \\ 
RelTransformer & WCE  & \textbf{19.2} & \textbf{20.0} & \textbf{19.5} & \textbf{15.7} & \textbf{13.6} &  \textbf{13.5}& \textbf{10.3} & \underline{\textbf{$\text{8.7}$}} & \underline{\textbf{$\text{8.1}$}} \\
\end{tabular}
}
\caption{Relationship triplet performance on GQA-LT dataset. SO = (subject, object), SR = (subject, relation) and OR = (object, relation). }
\label{compositional_result}
\end{table*}

\subsection{Further Analysis}
To analyze our results in more depth, we quantify our model's improvement per each class and visualize them in Fig \ref{gqalt_improvement}. We provide the contrast between RelTransformer and LSVRU with cross entropy loss. From the figure, we can observe that RelTransformer can improve the the majority of classes on both datasets. In particular, RelTransformer improves 173 relations while only worsening 32 ones on VG8K-LT dataset with most performance gain from the medium and few classes.

\begin{figure}
\centering
 \includegraphics[width=1.0\linewidth]
    {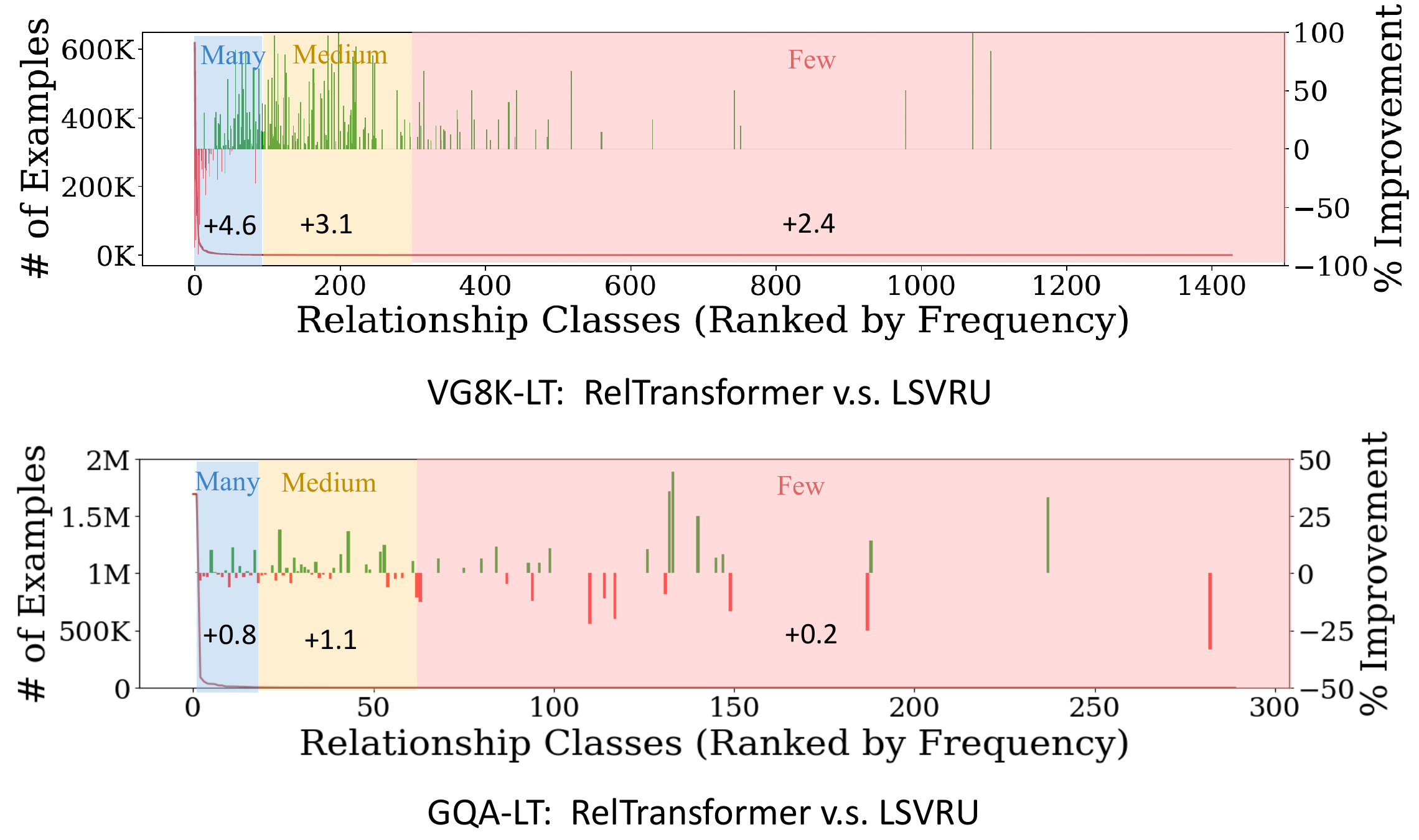}   
\caption{Per-class relationship accuracy comparisons between our RelTransformer and LSVRU \cite{lsvru} baseline on VG8K-LT(top) and GQA-LT(bottom) dataset. The green bars indicate the improvement of RelTransformer over LSVRU, red bars indicate worsening and no bars mean no change.  The left-side y-axis represents the number of examples per class. The right-side y-axis shows the absolute accuracy improvement. The x-axis represents the relation classes which are sorted by their frequency.}
\label{gqalt_improvement}
\end{figure}

\noindent\textbf{GQA-LT and VG8K-LT Compositional  Prediction.} The compositional prediction is the correct prediction of the subject, relation, and object together.
This could trigger a more skewed long-tail distribution due to its combinatorial nature. To evaluate our model's compositional behaviors, we follow the work \cite{ltvrd} to group the classification results by the pairs of (subject, object), (subject, relation), and (object, relation). The results are provided in Table \ref{compositional_result}, and we can see noticeable performance improvement on all the classes in contrast to the baselines. But we also observe that both RelTransformer(CE) and LSVRU (CE) perform better than the ones combining with focal loss and WCE in many categories, which differs from the results for only predicting the relations. The main reason is that those class imbalance losses hurt more ``head" performance on subject/object compared to CE (see supplementary) and they are reflected in the compositional prediction. 


\begin{figure}
\centering
 \includegraphics[width=1.0\linewidth]
    {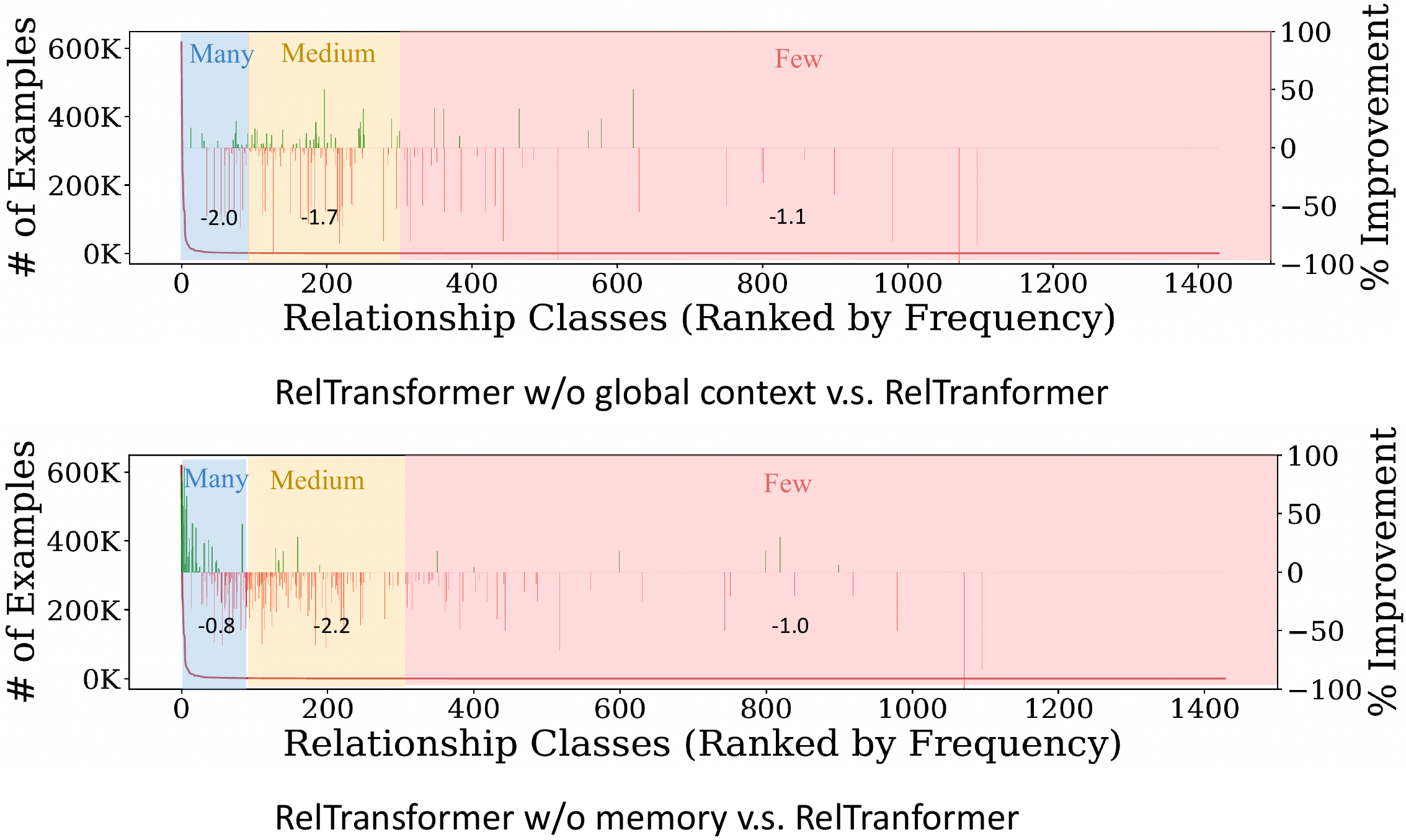}   
\caption{Per-class accuracy comparisons between the RelTransformer and its version without global-context encoder and memory attention using CE loss from Table \ref{ablation}. 
}
\label{gqalt_improvement}
\end{figure}

\begin{table}[t]
\centering
\small

\scalebox{0.9}{
\begin{tabular}{lcccccccc}
\multirow{2}{*}{Models} &\multirow{2}{*}{Losses} & many &  med &  few &  all \\
& &100 & 300 & 1,600 & 2,000\\
\toprule
\rowcolor{Gray}
Full model & CE & \textbf{26.8} & \textbf{18.6} & \textbf{15.0} & \textbf{16.1} \\
\xmark global & CE& 24.8 & 16.9 &13.9  & 14.8  \\
\xmark mem &CE&  26.0& 16.4& 14.0& 15.0 \\
\midrule
\rowcolor{Gray}
Full model & Focal Loss & \textbf{30.5} & \textbf{22.8}& \textbf{14.8} & \textbf{16.8} \\
\xmark global & Focal Loss&  28.9 & 19.8 & 13.3 & 15.1 \\
\xmark mem & Focal Loss& 30.1 &  20.8 & 14.2 & 15.9 \\
\midrule
\rowcolor{Gray}
Full model & DCPL & \textbf{37.3} & \textbf{27.6} & \textbf{16.5} & \textbf{19.2} \\
\xmark global & DCPL&  36.1 & 25.4 & 15.7& 18.2 \\
\xmark mem & DCPL& 37.0 &  26.8 & 16.1 & 18.8 \\
\midrule
\rowcolor{Gray}
Full model & WCE & \textbf{36.6} & \textbf{27.4} & \textbf{16.3} & \textbf{19.0} \\
\xmark global & WCE& 35.6 & 24.5& 15.2 & 17.6 \\
\xmark mem & WCE& 36.5 & 27.2 & 16.0& 18.7 
\end{tabular}}
\caption{Ablation study of RelTransformer on VG8K-LT dataset. global and mem represent the global-context encoder and memory attention module, respectively. \xmark represents the removal operation. Our default setting is marked in \colorbox{Gray}{gray}.}
\label{ablation}
\end{table}

\begin{figure*}[h!]
\centering
\scalebox{0.7}{
\begin{minipage}{0.33\textwidth}
 \includegraphics[width=1.0\linewidth,height =1.0\linewidth]
    {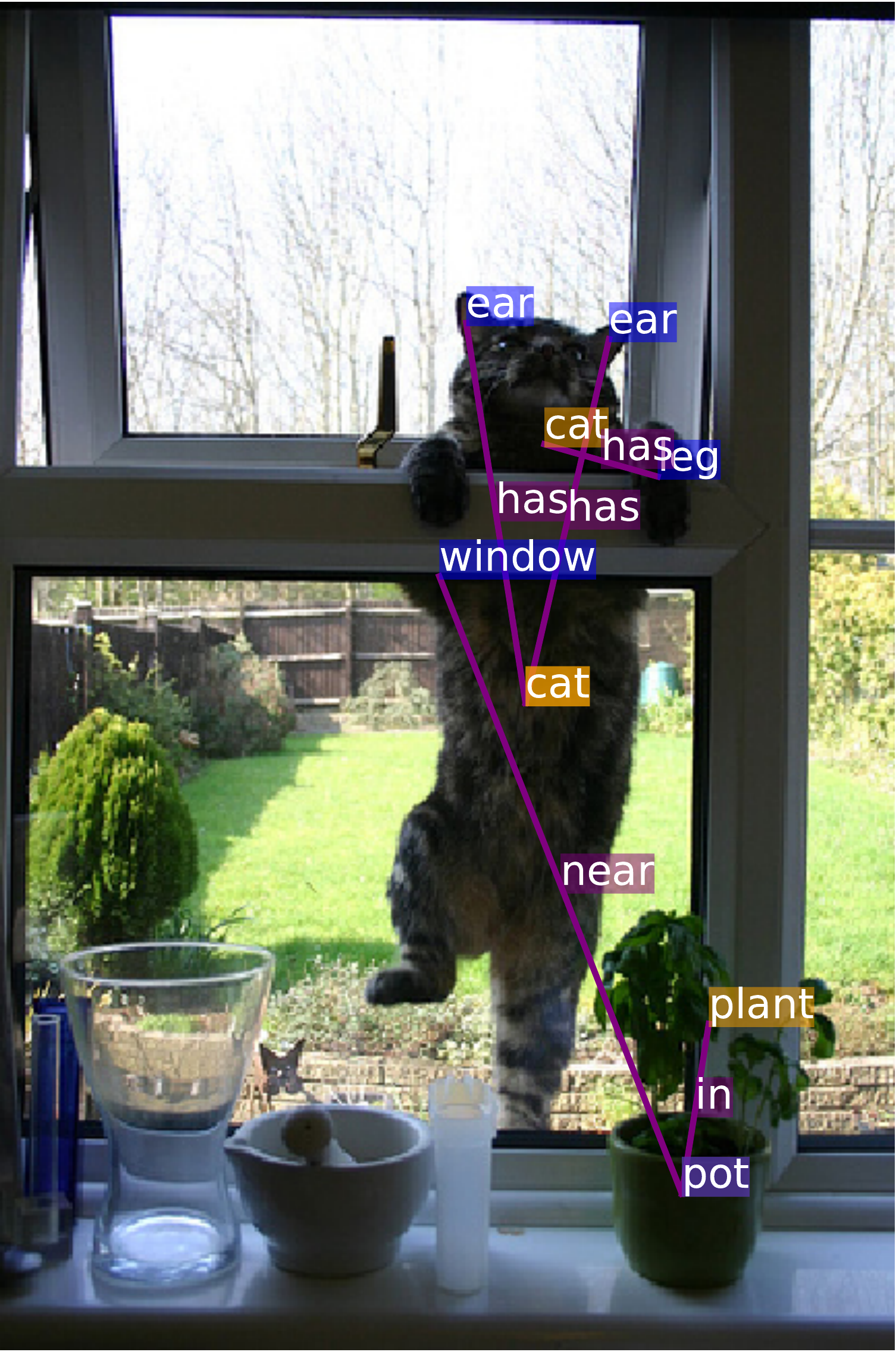}   
\end{minipage} \hfill
\begin{minipage}{0.33\textwidth}
\centering
 \includegraphics[width=1.0\linewidth,height =1.0\linewidth]
   {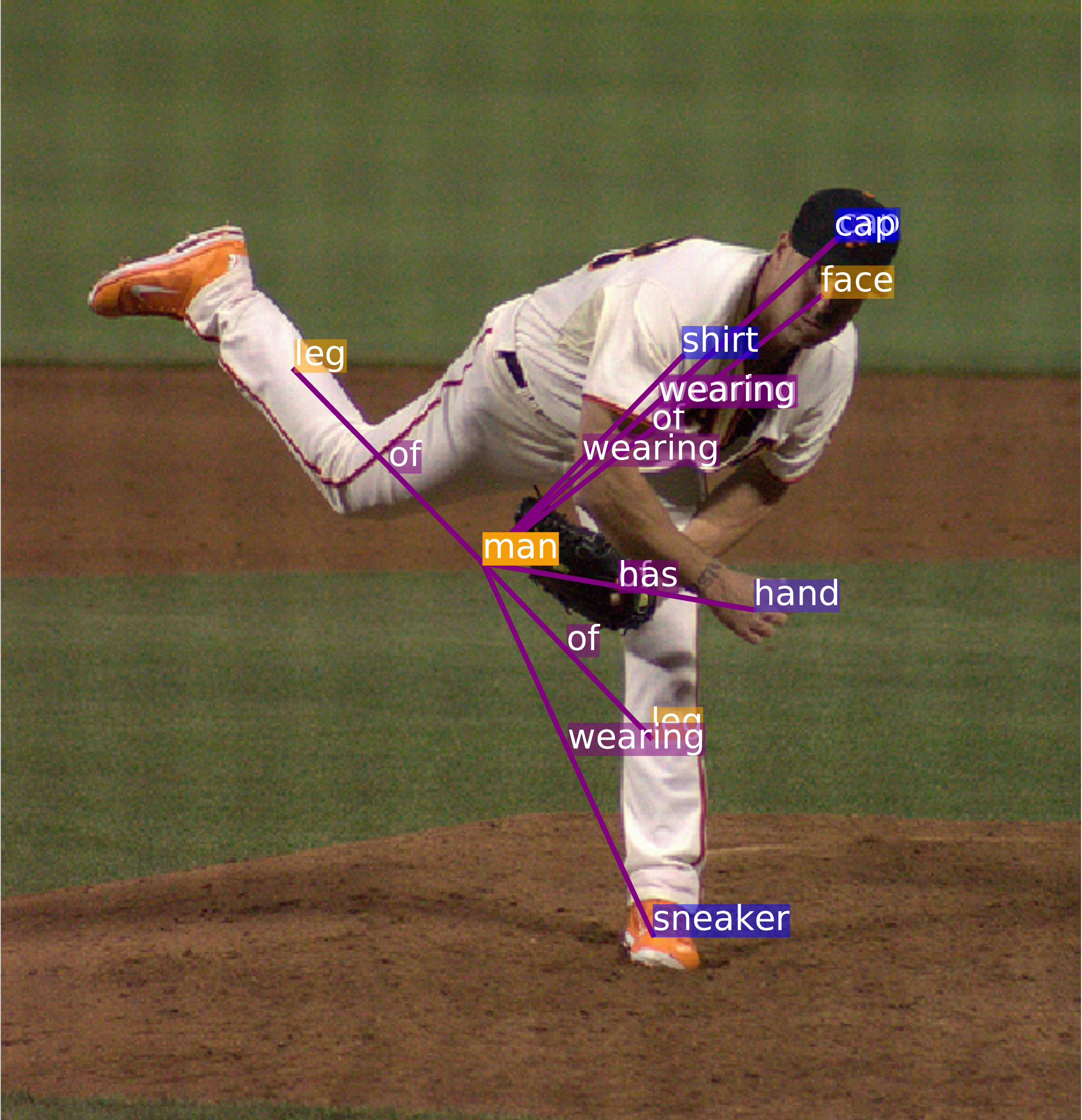}  
\end{minipage} \hfill
\begin{minipage}{0.33\textwidth}
\centering
 \includegraphics[width=1.0\linewidth,height =1.0\linewidth]
    {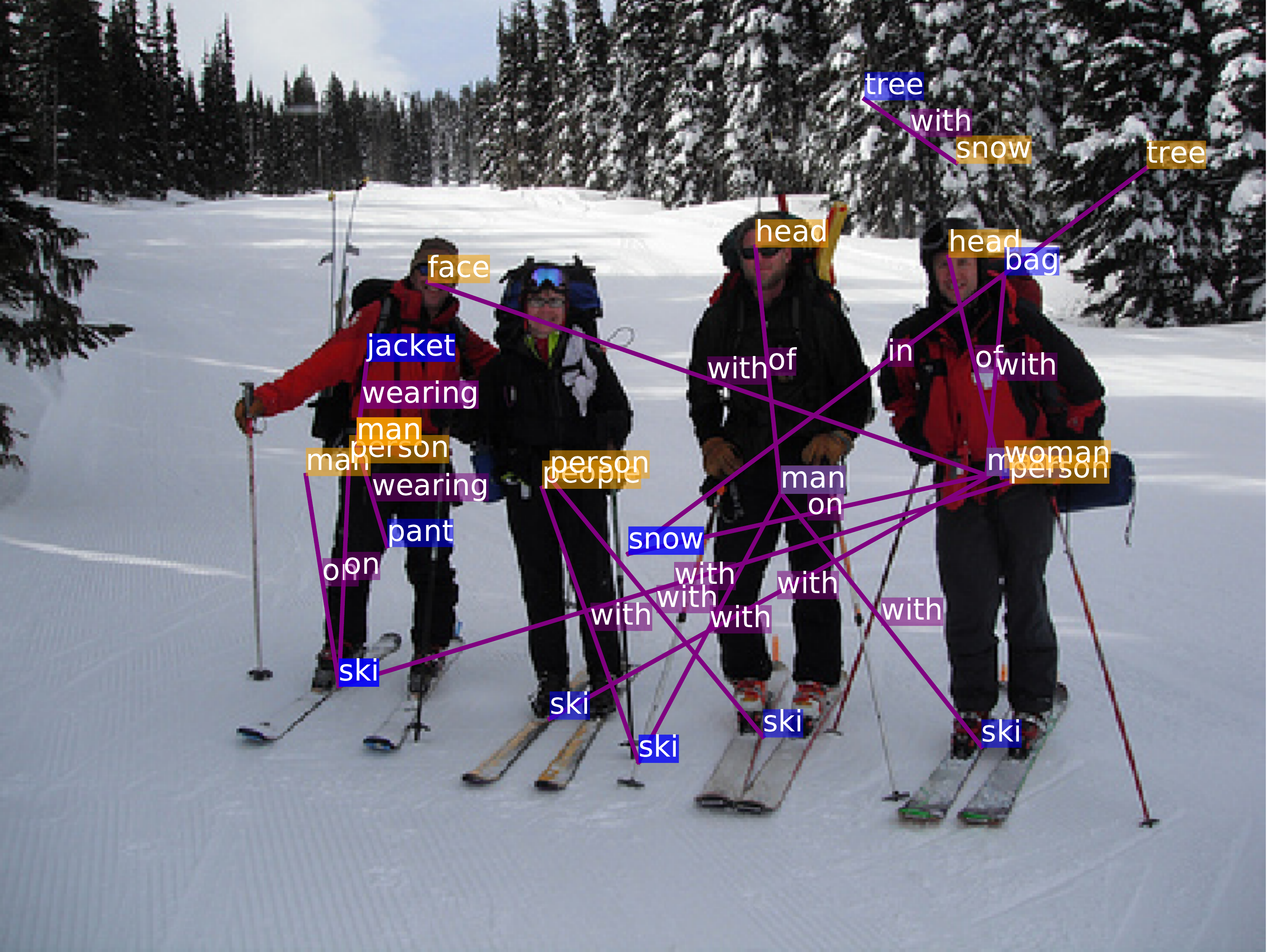}   
\end{minipage} \hfill}
\caption{The visualization of relation recognition with our model on VG200 dataset}
\label{VG200_examples}
\end{figure*}

\subsection{Ablation Studies}
To quantify the contributions of each component to the whole model performance, we ablate and evaluate our RelTrasnformer in different versions on VG8K-LT dataset as shown in Table \ref{ablation}. We choose VG8K-LT instead of GQA-lT since it is more challenging and covers more classes.


\noindent\textbf{The role of global-context encoder.} To investigate the effect of the global-context encoder, we ablate our RelTransformer with the version without learning global context. The results in Table \ref{ablation} indicate that the performance will drop on all the categories if we exclude it. It brings the performance down by 1.45\% accuracy (acc) on many, 2.35\% acc on medium, and 1.13\% acc on few when we average all the combined loss functions' results. This analytic shows that incorporating global context can benefit all categories with gaining the most performance on medium classes. We also demonstrate its per-class analysis in Fig. \ref{gqalt_improvement}, from which we can observe that the performance drops in most classes.

\noindent\textbf{The role of memory attention.} The persistent memory vectors are aimed to augment the relation representation with more useful ``out-of-context" information, especially for the low-frequent relations. In Table \ref{ablation}, we evaluate the version without the memory attention, and we can observe that it brings a performance reduction for all the categories. It drops the performance by 0.4\% acc on many, 1.2\% acc on medium, and 0.58\% acc on few relations when we average the results from all the loss functions. The medium and few are more influenced than many relations, indicating that memory attention can benefit the infrequent relation classes more. We also show the per-class analysis with removing the memory in Fig. \ref{gqalt_improvement}, where we can find the performances for medium and few relations drop most.


\subsection{Qualitative Results}
We randomly sample $2$ images which include long-tail relationships from the testing dataset. We evaluate RelTransformer and LSVRU \cite{lsvru} in terms of CE and WCE loss functions and contrast them in Fig. \ref{demonstrations}. It can be observed that LSVRU predicts very trivial relationships such as ``behind", ``under," or ``to the left of". Those trivial relationships convey a vague and inaccurate understanding of the image content. Our RelTransformer instead can describe the relations more accurately. e.g., it predicts ``man sniffing wine" for the bottom image. ``sniffing" is a long-tail relation type and this prediction exactly matches the ground truth. We also randomly sample $3$ images from VG200 dataset and visualize our prediction results in Fig. \ref{VG200_examples}.

\section{Conclusion}
We presented a Transformer-based long-tail visual relationship recognition model, dubbed RelTransformer, which directly connects the relations with all the visual objects via the attention mechanism. Empirically, we organize the whole scene into the relation-triplet and global-scene context, and attentively aggregate their information to the relation representation under our message-passing flow. We also propose a memory attention module to augment the relation representation with ``out-of-context" information, which is shown to be more effective for infrequent relations. 
With our design, RelTransformer surpasses all previous stat-of-the-art results on GQA-LT, VG8K-lT, and VG200 datasets.





{\small
\bibliographystyle{ieee_fullname}
\bibliography{egbib}
}

\end{document}


\title{RelTransformer: A Transformer-Based Long-Tail Visual Relationship Recognition Supplementary}  

\author{
  Jun Chen\textsuperscript{\rm 1},   Aniket Agarwal\textsuperscript{\rm 2}, Sherif Abdelkarim\textsuperscript{\rm 1},
  Deyao Zhu\textsuperscript{\rm 1},  Mohamed Elhoseiny\textsuperscript{\rm 1}\\
  \textsuperscript{\rm 1}{King Abdullah University of Science and Technology} \\ 
\textsuperscript{\rm 2}{ Indian Institute of Technology} \\
\texttt{\{jun.chen,deyao.zhu,mohamed.elhoseiny\}@kaust.edu.sa} \\
 \texttt{aagarwal@ma.iitr.ac.in}, \texttt{sherif.abdelkarim91@gmail.com} 
}

\maketitle

\appendix

\section{Training and Implementation Details}
To have fair comparisons with all the baseline models, we follow the same experimental setup as \cite{ltvrd} for GQA-LT and VG8K-LT evaluation and also the same setup as \cite{lsvru} for VG200 evaluation. We train our model and all its ablations with 8 V100 GPUs. The batch size is 8. We train our models with 12, 8, and 7 epochs on GQA-LT, VG8K-LT, and VG200 datasets, respectively. The hidden size of $h$ is 768. The number of Transformer heads is 12. Relational and global-context encoders both have 2 layers. The memory size is 100 $\times$ 768. We use the Faster R-CNN \cite{ren2015faster} with VGG-16 \cite{vgg} backbone to extract the object proposal features. We also apply the pretrained Word2Vec \cite{word2vec} embeddings to represent the relation and object labels. 
\begin{figure}[h!]
\centering
 \includegraphics[width=0.8\linewidth]{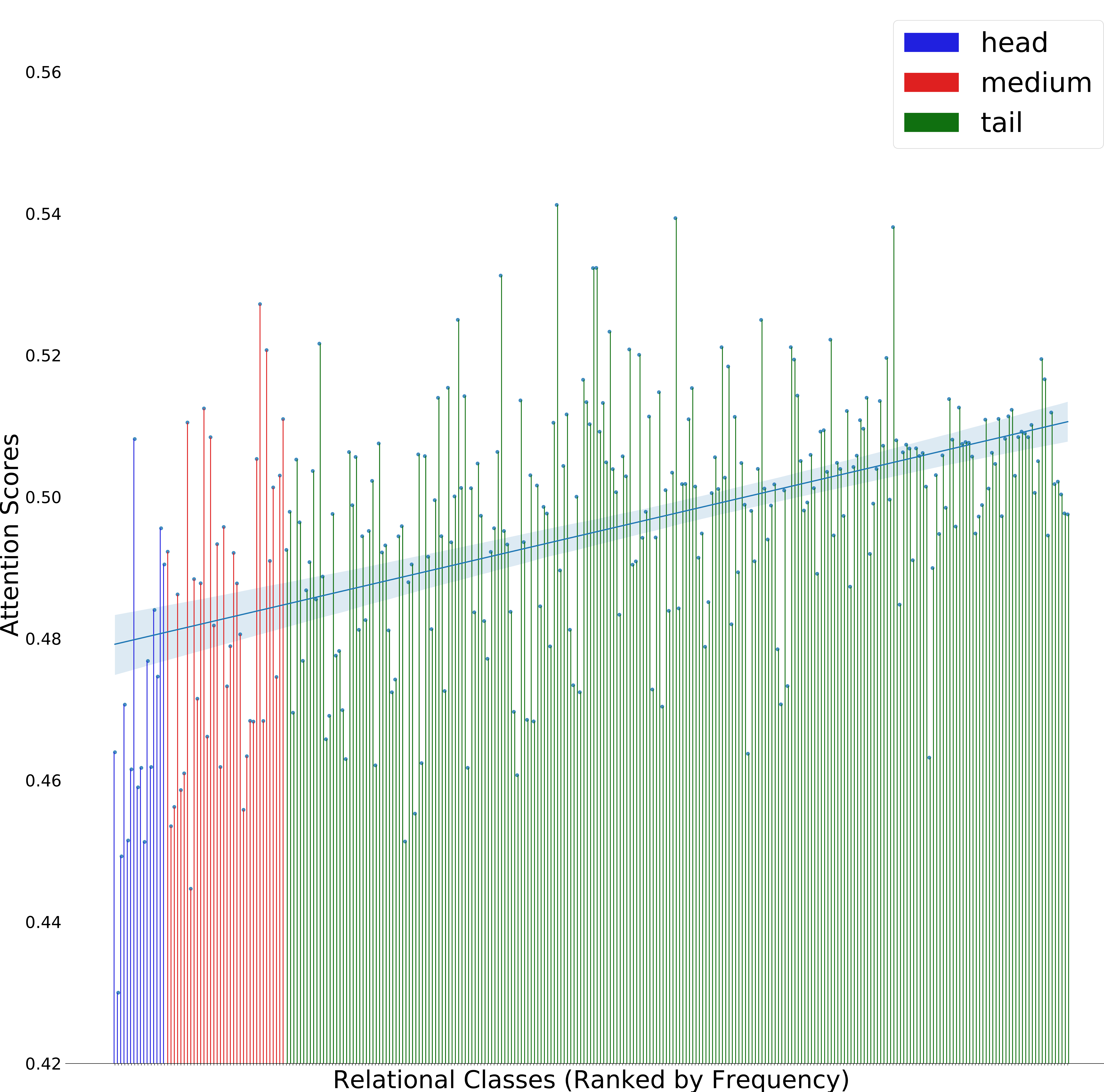}
 \caption{Average memory attention scores per each relation class on evaluation of GQA-LT. The relation classes are ranked by its frequency (from high to low)}
\label{mem_attention}
\end{figure}

\section{Subject and Object Per-Class Accuracy}
We also provide the per-class accuracy for subjects and objects in Table \ref{sub_obj_results}. We can observe that our RelTransformer can also easily outperform many baselines. Combining RelTransformer with CE loss can outperform all the baseline models on each category and it significantly improves over the ``many" category. Combining it with WCE can furthermore improve over ``medium" and ``tail" categories on both datasets. These results demonstrate the effectiveness of our model on the subject and object prediction. We also notice that RelTransformer (WCE or DCPL) drops the performance over ``many" categories compared to CE loss function; We hypothesis that it is due to the assigned lower weights on the high-frequent subjects/objects, and lower weights will lead to the lower confidence values during the classification. This phenomenon can explain why RelTansformer (WCE) underperforms on the ``many" classes and why it underperforms on the compositional prediction for ``many" and ``medium" classes.

\begin{table*}[h]
\centering
\begin{tabular}{ ll| cccc|cccc}
 \multicolumn{2}{l}{} & \multicolumn{4}{c}{VG8K-LT} & \multicolumn{4}{c}{GQA-LT} \\
 \multirow{2}{*}{Architecture} & \multirow{2}{*}{Learning Methods} & many &  medium &  few &  all &  many &  medium &  few &  all \\
  &   &  267 & 799 &  4,264 &  5,330 &  86 &  255 &  1,362 & 1,703 \\

\toprule
LSVRU & VilHub \cite{ltvrd} & 61.6 & 20.3 & 10.1 & 14.2 & 68.6 & 44.0 & 10.3 & 18.3 \\
LSVRU & VilHub + RelMix \cite{ltvrd}  & 59.5 &15.1 & 10.4& 13.6 & 68.8 &42.1& 10.1 & 18.1 \\
LSVRU & OLTR \cite{oltr} & 56.8 & 12.0 & 9.6 & 12.3 & 68.2 & 37.2 & 7.0 & 14.6 \\
LSVRU & EQL \cite{eql} & 56.9 & 12.1 & 10.0 & 12.7 & 68.9 & 43.7 & 10.0 & 18.0\\
LSVRU & $\text{Counterfactual} ^{\sharp}$\cite{tang2020unbiased} & 57.3 & 11.1 & 8.5 & 11.4 & 68.3 & 37.0 & 6.9 & 14.5\\
\midrule
LSVRU & CE & 57.3 & 11.1 & 8.5 & 11.4 & 68.3 & 37.0 & 6.9 & 14.5\\
\rowcolor{Gray2}
RelTransformer & CE & \textbf{\underline{67.1}} & \textbf{25.9} & \textbf{11.5} & \textbf{16.5} &\underline{\textbf{78.0}} & \underline{\textbf{56.6}} & \textbf{14.2} & \textbf{23.8} \\ 
\midrule
LSVRU & Focal Loss \cite{focalLoss} & 58.1 & 13.9 & 8.9 & 12.1 & 68.2 & 39.2 & 7.5 & 15.3 \\ \rowcolor{Gray2}
RelTransformer & Focal Loss & \textbf{65.6} & \textbf{21.7} & \textbf{10.8} & \textbf{15.2} & \textbf{75.0} & \textbf{51.4} & \textbf{11.9} & \textbf{21.0}\\
\midrule
LSVRU & DCPL\cite{decoupling} & \textbf{53.8} & 5.9 & 7.9 & 9.9 & \textbf{64.0} & 35.3 & 6.4 & 13.7\\
\rowcolor{Gray2}
RelTransformer & DCPL & 50.3 & \textbf{30.9} & \textbf{13.4}& \textbf{17.8} & 51.8 & \textbf{44.6} & \textbf{19.2} & \textbf{24.7}\\
\midrule
LSVRU & WCE& \textbf{52.8} & 27.2 & 10.8 & 14.5 & \textbf{53.4} & 42.0 & 14.0 & 20.2 \\ 
\rowcolor{Gray2}
RelTransformer& WCE & 50.1& \underline{\textbf{31.3}}& \underline{\textbf{13.7}} &\underline{\textbf{18.0}} & 50.3& \textbf{46.2}&\underline{\textbf{28.7}}& \underline{\textbf{32.4}} \\
\end{tabular}
\caption{Average per-class accuracy for subject/object. We separately evaluate the average  per-class  accuracy for many, medium, few, and all categories. The best performance from each category is underlined. $\sharp$ denotes our reproduction. Our model is denoted in \colorbox{Gray2}{gray}}
\label{sub_obj_results}
\end{table*}

\begin{table*}[h]
\centering
\begin{tabular}{ ll| cccc|cccc}

 \multicolumn{2}{l}{} & \multicolumn{2}{c}{Per-example Accuracy} \\
 Architecture & Learning Methods & Subject/Object & Relation \\
\toprule
LSVRU & CE & 51.9 & 94.8\\
LSVRU & VilHub \cite{ltvrd} &53.9 & 91.2 \\
LSVRU & VilHub + RelMix \cite{ltvrd}  & 53.5 &91.0  \\
LSVRU & EQL \cite{eql} & 51.1 & 93.9\\
LSVRU & WCE& 37.6 & 72.6  \\ 
\rowcolor{Gray2}
RelTransformer & CE & \textbf{\underline{62.5}} & \textbf{\underline{95.1}} \\
\rowcolor{Gray2}
RelTransformer & Focal Loss \cite{focalLoss} & 59.4 & 95.0\\
\rowcolor{Gray2}
RelTransformer & DCPL \cite{decoupling} &34.3 & 80.5 \\
\rowcolor{Gray2}
RelTransformer& WCE & 34.2& 74.2 \\
\end{tabular}
\caption{Per-example Accuracy on GQA-LT. The best performance from each category is underlined. Our model is denoted in \colorbox{Gray2}{gray}}
\label{per_example_acc}
\end{table*}

\section{Per-Example Accuracy}
We provide the per-example accuracy for the subjects and objects on GQA-LT dataset. It shows that RelTransformer (CE) achieves the best performance on predicting subjects/objects and relations among all the baselines. Per-example accuracy is mainly dominated by the ``head" classes since they are very large in example numbers. This indicates that RelTransformer (CE) improves both head and tail classes. Compared to the best baseline LSVRU (CE), RelTransformer (CE) improves it by 10.6 acc on subject/object and 0.3 acc on relation predictions. However, we could also notice that RelTransformer (WCE or DCPL) brings the performance down on per-example accuracy, and this phenomenon is also observed in LSVRU baselines. But the results in Table \ref{sub_obj_results}(Supplementary) and Table 1 (main paper) shows that RelTransformer (WCE or DCPL) has a good-performing result on ``medium" and ``tail" categories. This indicates that combining with class-imbalance loss functions can benefit low-frequent class predictions with the cost of the performance from a few number of top-frequent classes; RelTransformer improves the results more general.



\section{Memory Attention Further Analysis}

To further investigate the role of memory attention, we compute the memory attention scores, $J-\alpha$ in our fusion function Eq. \ref{fusion}, for each testing example from well-trained RelTransformer (WCE) on the GQA-LT dataset. We average the attention scores per each relation-class and demonstrate them in Fig. \ref{mem_attention}. The relation classes are ranked according to their frequency in the training data. We can observe a clear increasing trend for the attention scores from high-frequent relations to low-frequent ones, meaning that the features from memory can contribute to the long-tail relations more than the head-relations, which can reflect why memory can gain more improvement on the ``medium" and ``tail" classes. 
\begin{equation}
\begin{split}
   g(x,y) &=  \alpha \odot x  + (J-\alpha) \odot y\\
   \alpha &= \sigma( W[x ; y] + b)
\end{split}
\label{fusion}
\end{equation}
 where $W$ is a 2D $\times$ D matrix. $b$ is a bias term. $[;]$ denotes the concatenation. $\odot$ is the Hadamard product. $J$ is an all-one matrix with the same dimensions as $\alpha$.



\section{Additional Qualitative Examples}
\begin{figure*}[h!]

\centering
\begin{minipage}{0.33\textwidth}
\centering
 \includegraphics[width=1.0\linewidth,height=1.0\columnwidth]
    {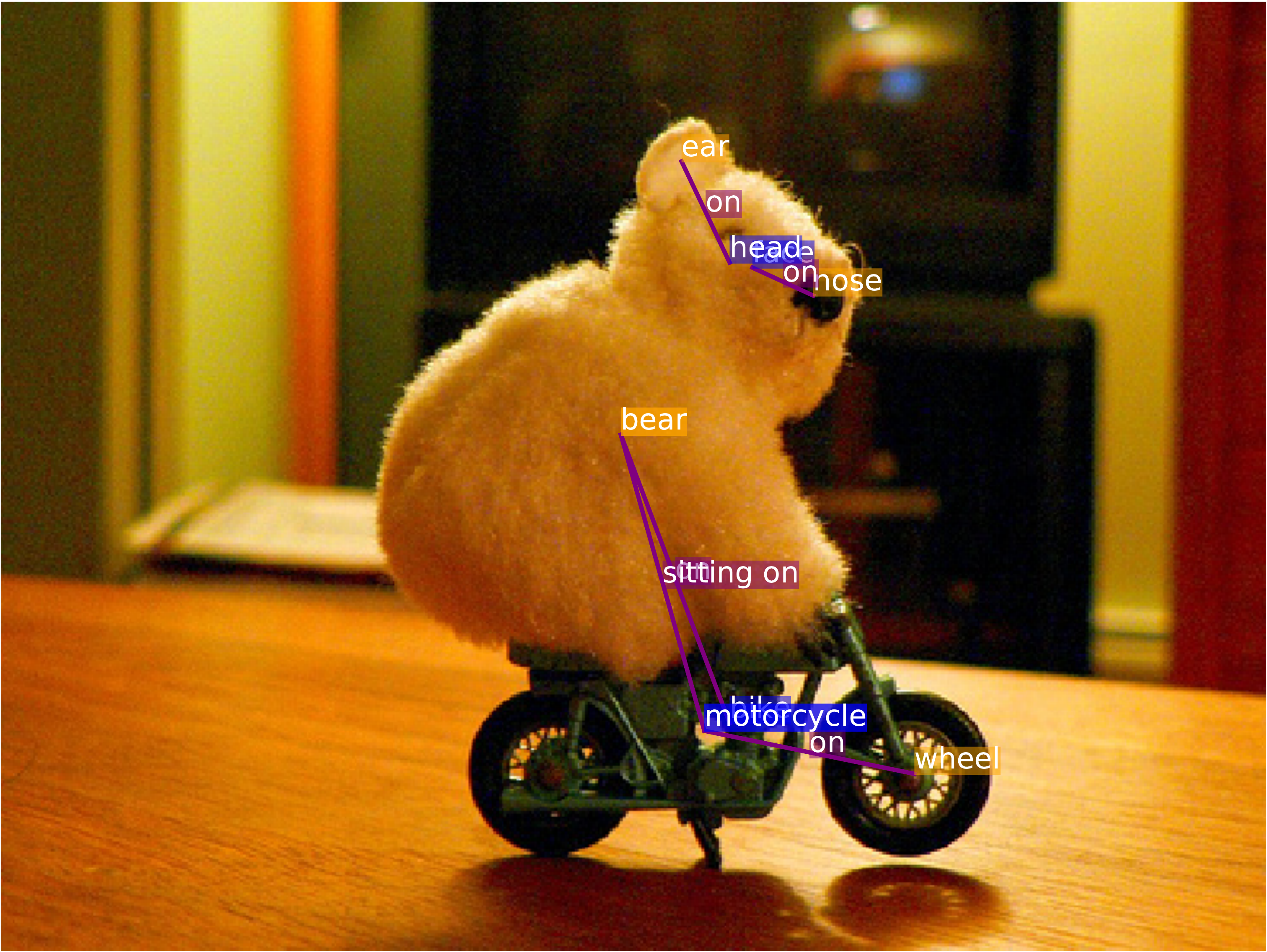}   
\end{minipage} \hfill
\begin{minipage}{0.33\textwidth}
\centering
 \includegraphics[width=1.0\linewidth,height=1.0\columnwidth]
   {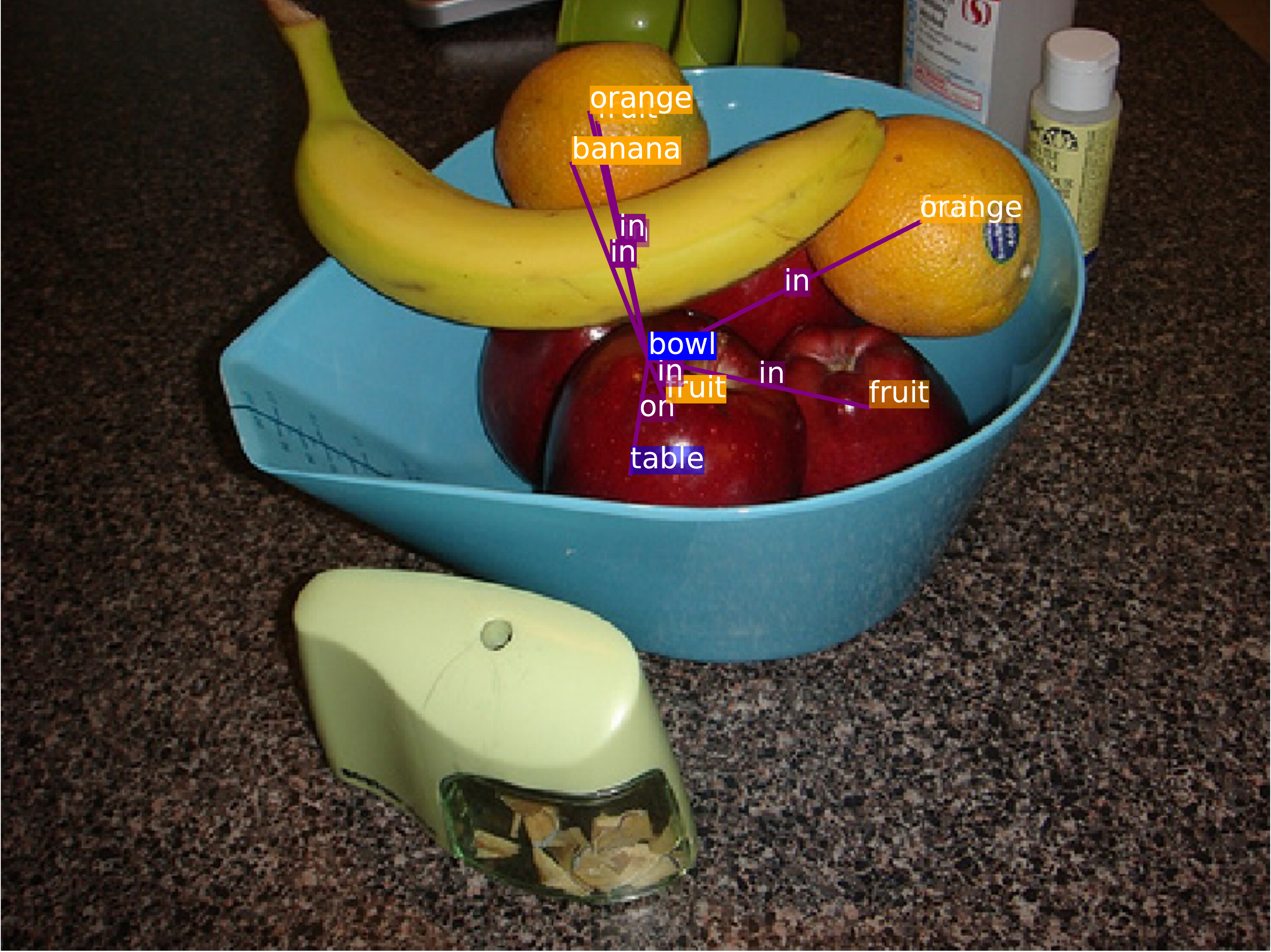}  
\end{minipage} \hfill
\begin{minipage}{0.33\textwidth}
\centering
 \includegraphics[width=1.0\linewidth,height=1.0\columnwidth]
    {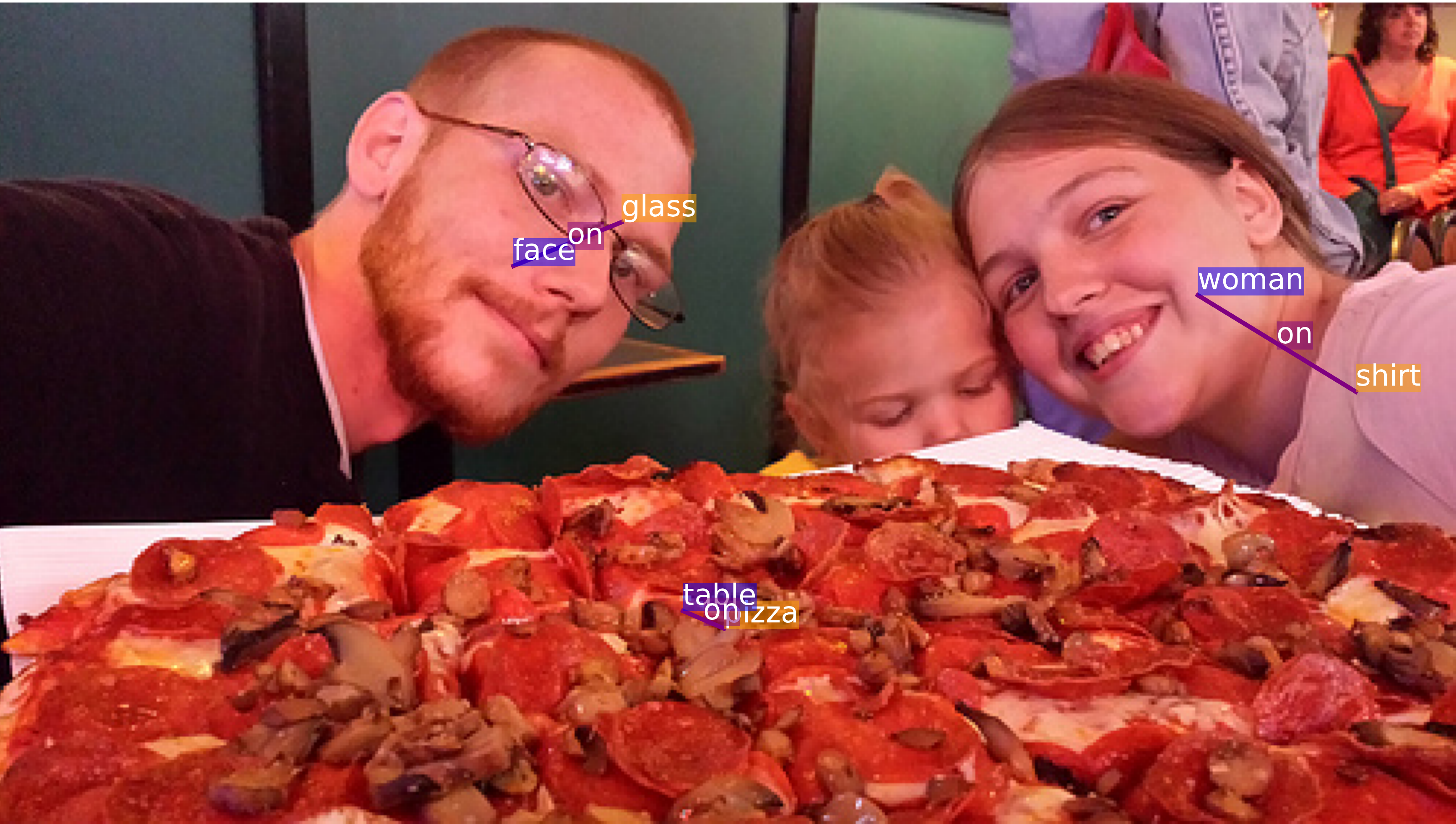}   
\end{minipage} \hfill
\begin{minipage}{0.33\textwidth}
\centering
 \includegraphics[width=1.0\linewidth,height=1.0\columnwidth]
   {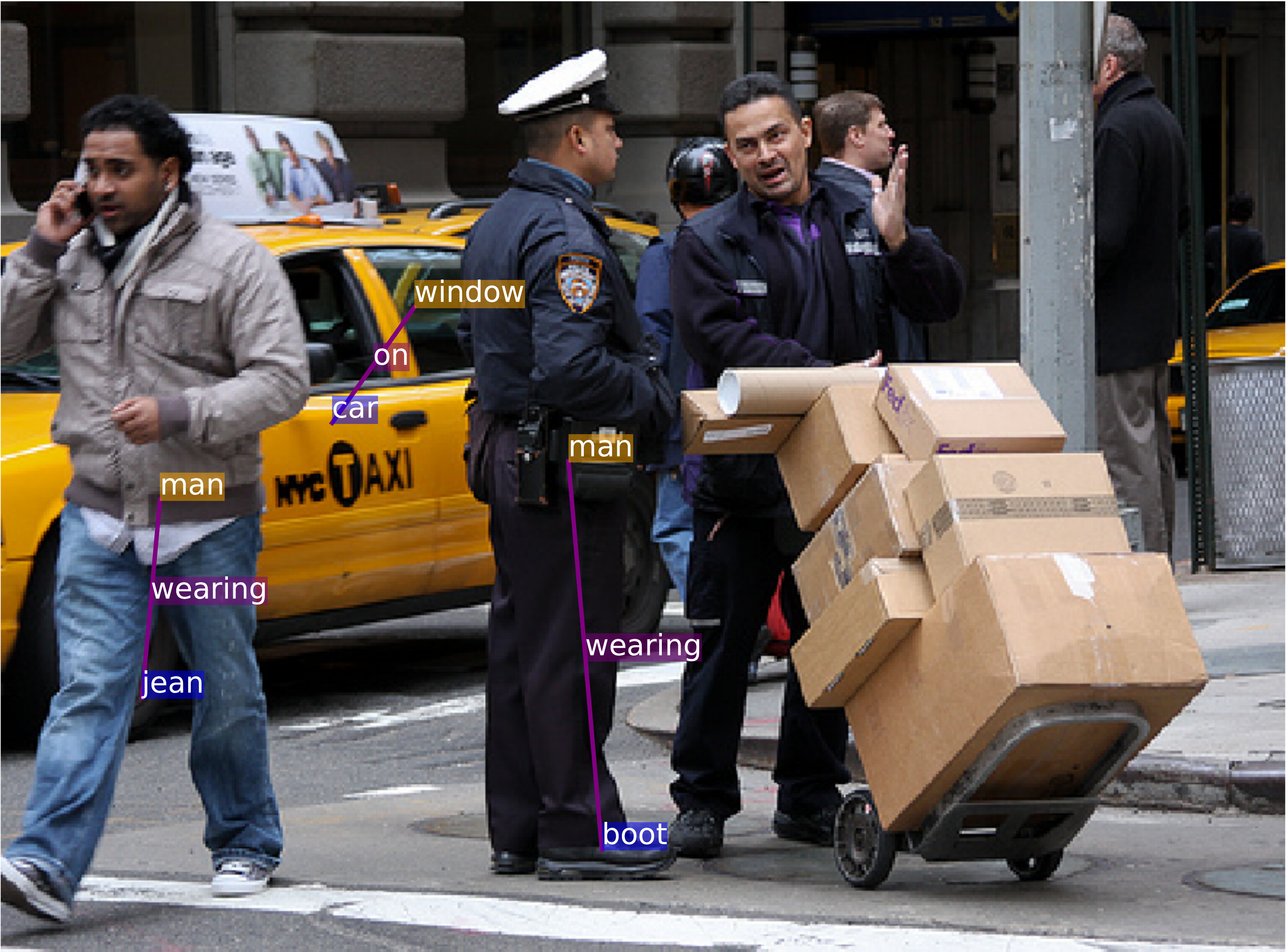}  
\end{minipage} \hfill
\begin{minipage}{0.33\textwidth}
\centering
 \includegraphics[width=1.0\linewidth,height=1.0\columnwidth]
    {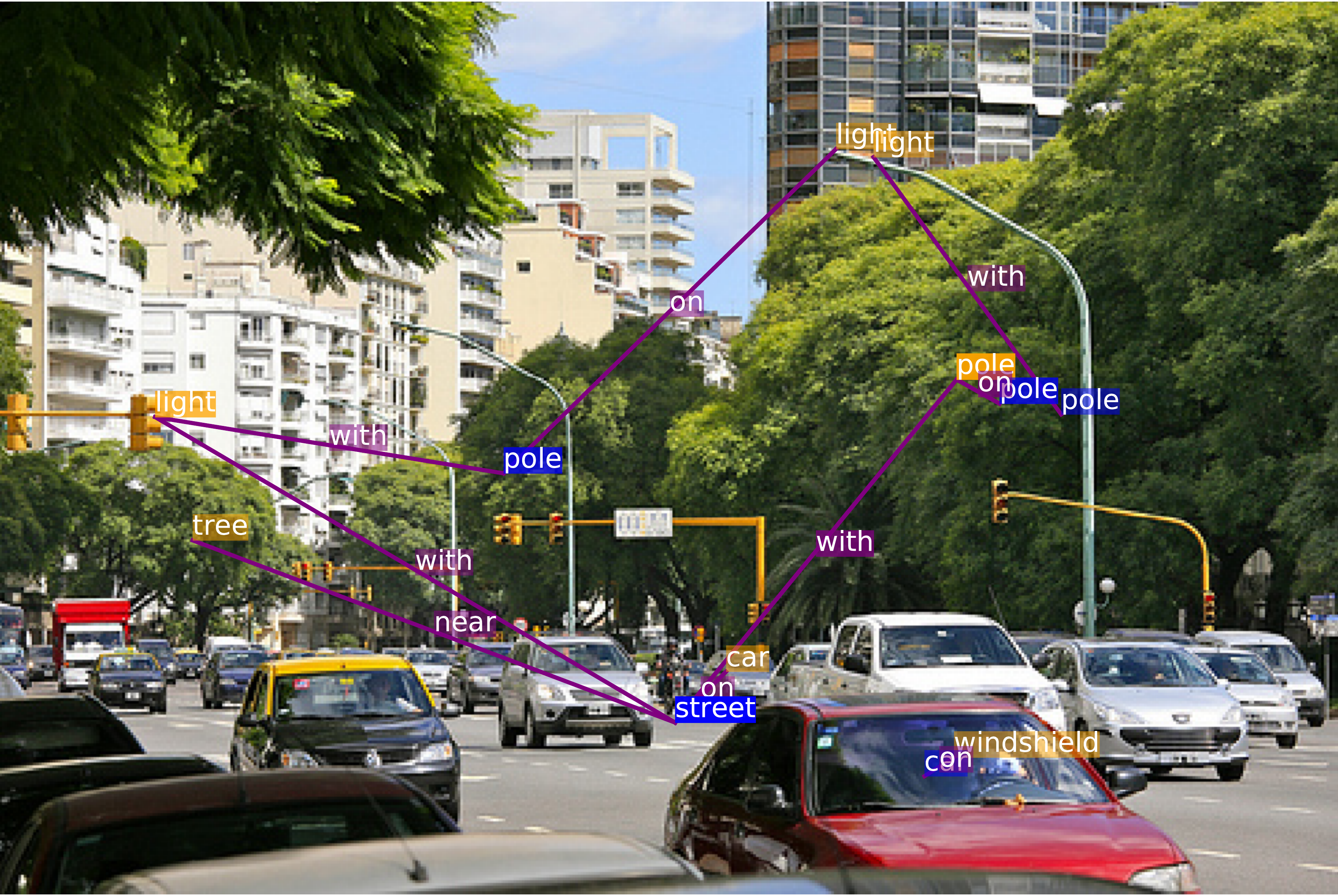}   
\end{minipage} \hfill
\begin{minipage}{0.33\textwidth}
\centering
 \includegraphics[width=1.0\linewidth,height=1.0\columnwidth]
   {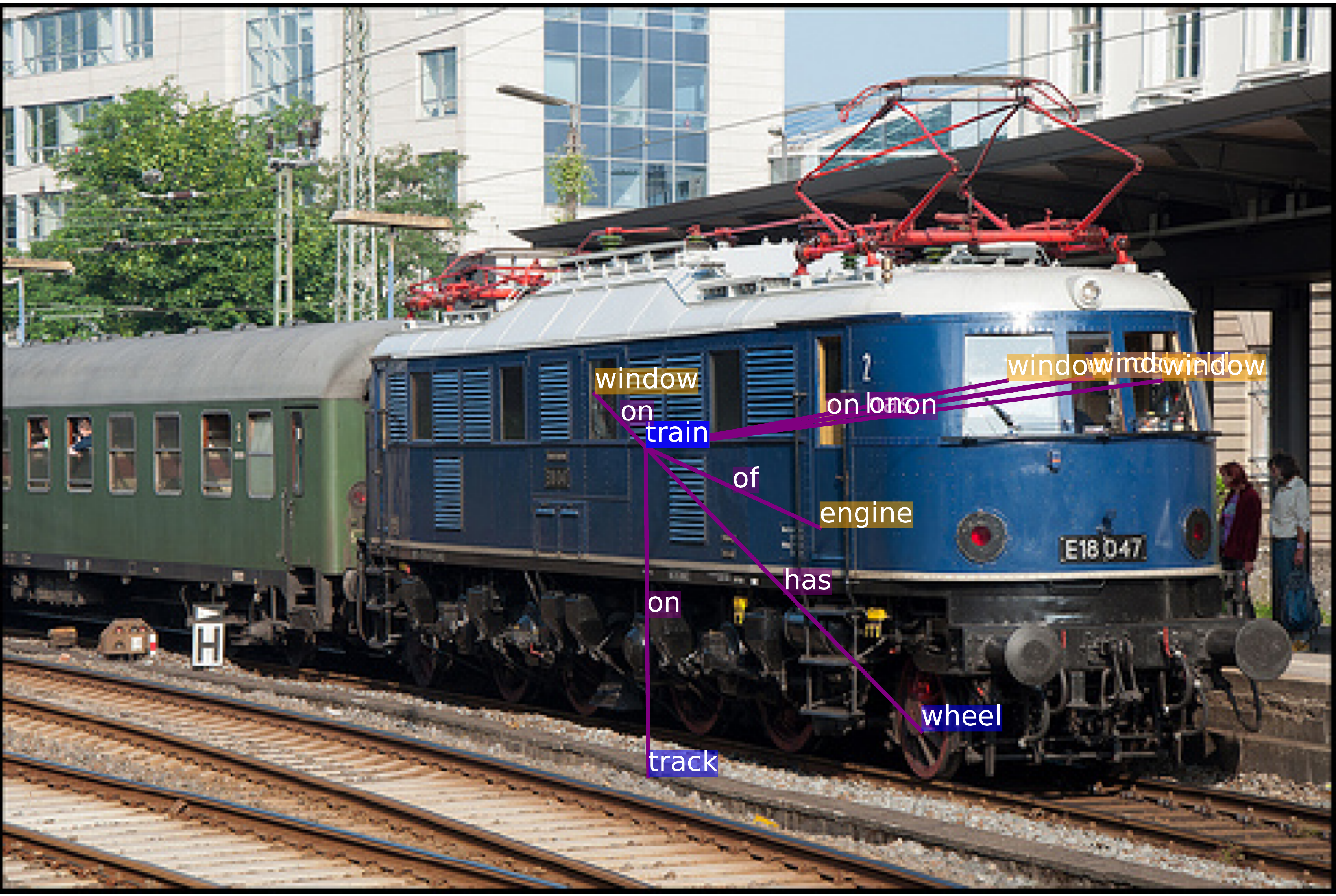}  
\end{minipage} \hfill
\begin{minipage}{0.33\textwidth}
\centering
 \includegraphics[width=1.0\linewidth,height=1.0\columnwidth]
    {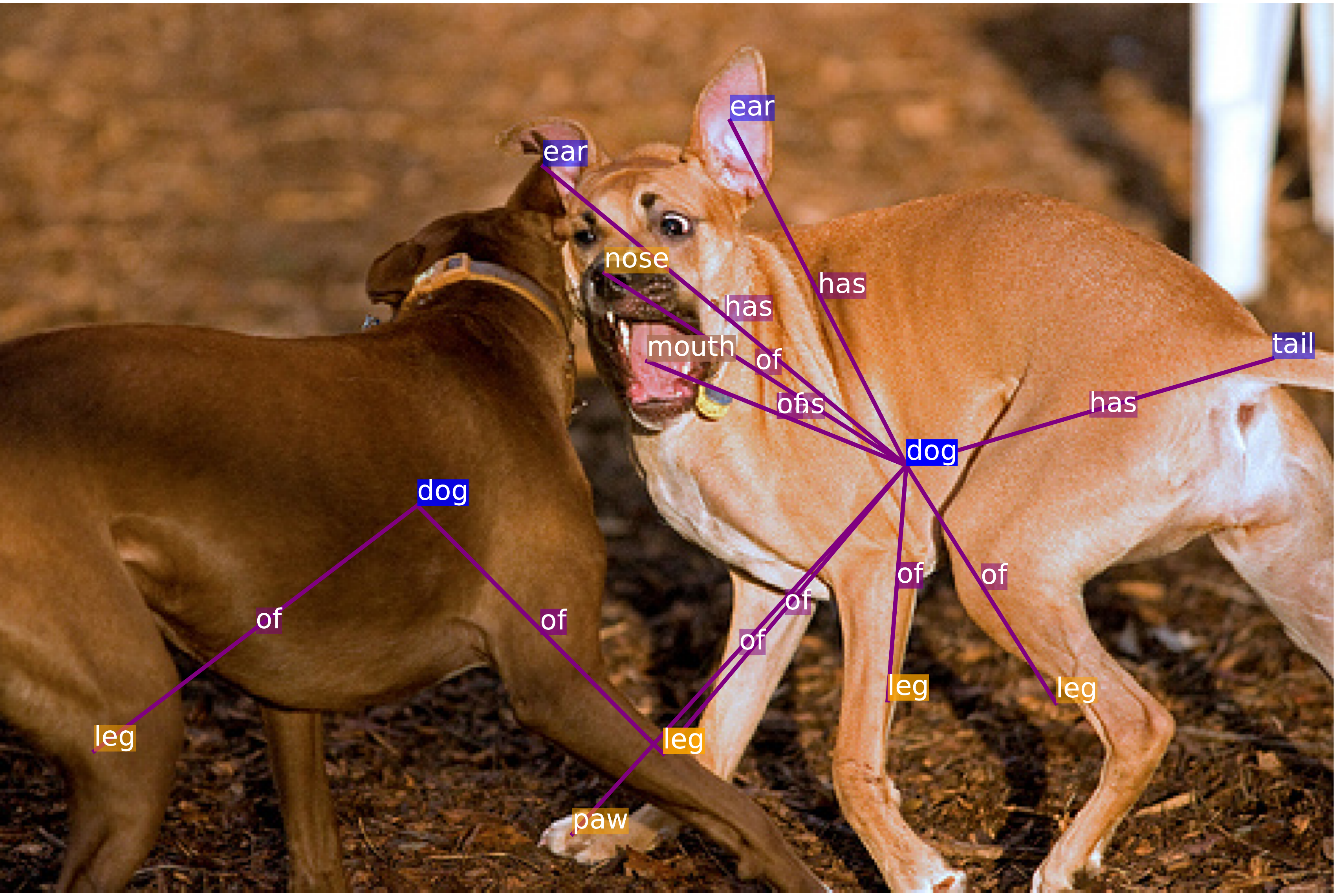}   
\end{minipage} \hfill
\begin{minipage}{0.33\textwidth}
\centering
 \includegraphics[width=1.0\linewidth,height=1.0\columnwidth]
   {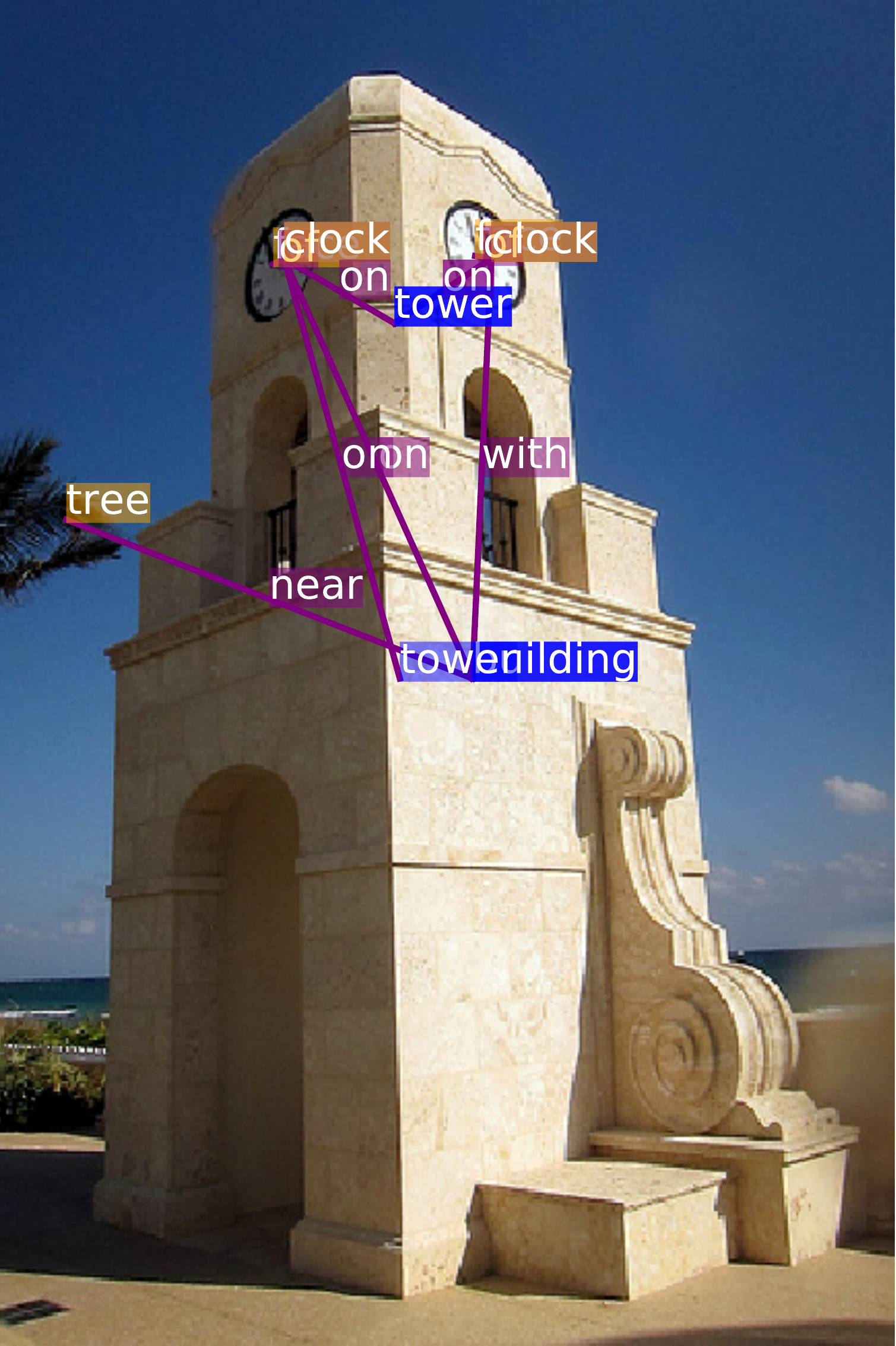}  
\end{minipage} \hfill
\begin{minipage}{0.33\textwidth}
\centering
 \includegraphics[width=1.0\linewidth,height=1.0\columnwidth]
    {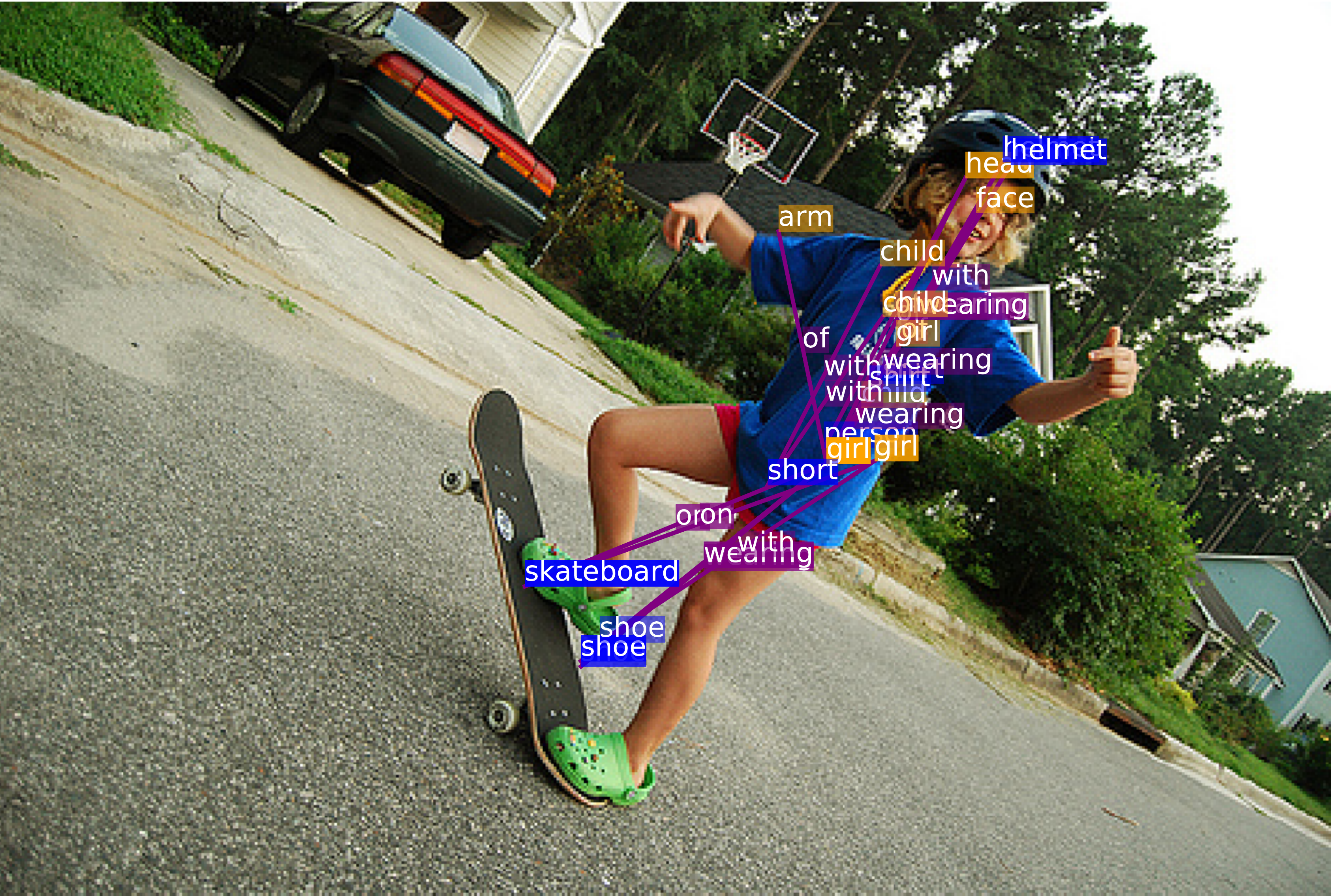}   
\end{minipage} \hfill
\caption{
Qualitative results on VG200 dataset}
\label{VG200_examples}

\end{figure*}

We show the relation prediction results on VG200 dataset in Fig. \ref{VG200_examples} and provide more long-tail relation prediction results on GQA-LT and VG8K-LT dataset in Fig. \ref{gqalt_examples} and \ref{vg8k_examples}. 


\begin{figure*}[h!]

\begin{minipage}{0.5\textwidth}
\centering
 \includegraphics[width=1.0\linewidth]
    {vg8k_figure/fig_1a_crop.pdf}   
\end{minipage} \hfill
\begin{minipage}{0.5\textwidth}
\centering
 \includegraphics[width=1.0\linewidth]
    {vg8k_figure/fig_1b_crop.pdf}
\end{minipage} \hfill

\begin{minipage}{0.5\textwidth}
\centering
 \includegraphics[width=1.0\linewidth]
    {vg8k_figure/fig_2a_crop.pdf}
\end{minipage} \hfill
\begin{minipage}{0.5\textwidth}
\centering
 \includegraphics[width=1.0\linewidth]
    {vg8k_figure/fig_2b_crop.pdf}
\end{minipage} \hfill

\begin{minipage}{0.5\textwidth}
\centering
 \includegraphics[width=1.0\linewidth]
    {vg8k_figure/fig_3a_crop.pdf}
\end{minipage} \hfill
\begin{minipage}{0.5\textwidth}
\centering
 \includegraphics[width=1.0\linewidth]
    {vg8k_figure/fig_3b_crop.pdf}
\end{minipage} \hfill

\begin{minipage}{0.5\textwidth}
\centering
 \includegraphics[width=1.0\linewidth]
    {vg8k_figure/fig_4a_crop.pdf} 
\end{minipage} \hfill
\begin{minipage}{0.5\textwidth}
\centering
 \includegraphics[width=1.0\linewidth]
    {vg8k_figure/fig_4b_crop.pdf}  
\end{minipage} \hfill

\begin{minipage}{0.5\textwidth}
\centering
 \includegraphics[width=1.0\linewidth]
    {vg8k_figure/fig_5a_crop.pdf}
    \title{ \hspace{20pt}  LSVRU  \hspace{70pt}  RelTransformer }
\end{minipage} \hfill
\begin{minipage}{0.5\textwidth}
\centering
 \includegraphics[width=1.0\linewidth]
    {vg8k_figure/fig_5b_crop.pdf} 
    \title{  LSVRU (WCE)  \hspace{50pt}      RelTransformer (WCE) }
\end{minipage} \hfill

\caption{
More long-tail relation prediction examples on GQA-LT dataset}
\label{gqalt_examples}

\end{figure*}


\begin{figure*}

\begin{minipage}{0.5\textwidth}
\centering
 \includegraphics[width=1.0\linewidth]
    {vg8k_figure/fig_6a_crop.pdf}
\end{minipage} \hfill
\begin{minipage}{0.5\textwidth}
\centering
 \includegraphics[width=1.0\linewidth]
    {vg8k_figure/fig_6b_crop.pdf}
\end{minipage} \hfill

\begin{minipage}{0.5\textwidth}
\centering
 \includegraphics[width=1.0\linewidth]
    {vg8k_figure/fig_7a_crop.pdf}
\end{minipage} \hfill
\begin{minipage}{0.5\textwidth}
\centering
 \includegraphics[width=1.0\linewidth]
    {vg8k_figure/fig_7b_crop.pdf} 
\end{minipage} \hfill

\begin{minipage}{0.5\textwidth}
\centering
 \includegraphics[width=1.0\linewidth]
    {vg8k_figure/fig_8a_crop.pdf}
\end{minipage} \hfill
\begin{minipage}{0.5\textwidth}
\centering
 \includegraphics[width=1.0\linewidth]
    {vg8k_figure/fig_8b_crop.pdf}  
\end{minipage} \hfill

\begin{minipage}{0.5\textwidth}
\centering
 \includegraphics[width=1.0\linewidth]
    {vg8k_figure/fig_9a_crop.pdf}
\end{minipage} \hfill
\begin{minipage}{0.5\textwidth}
\centering
 \includegraphics[width=1.0\linewidth]
    {vg8k_figure/fig_9b_crop.pdf}
\end{minipage} \hfill

\begin{minipage}{0.5\textwidth}
\centering
 \includegraphics[width=1.0\linewidth]
    {vg8k_figure/fig_10a_crop.pdf}
    \title{ \hspace{20pt}  LSVRU  \hspace{70pt}  RelTransformer }
\end{minipage} \hfill
\begin{minipage}{0.5\textwidth}
\centering
 \includegraphics[width=1.0\linewidth]
    {vg8k_figure/fig_10b_crop.pdf} 
    \title{  LSVRU (WCE)  \hspace{50pt}      RelTransformer (WCE) }
\end{minipage} \hfill



\caption{
More long-tail relation prediction examples on VG8K-LT dataset}
\label{vg8k_examples}

\end{figure*}

\newpage
\clearpage
\clearpage

{\small
\bibliographystyle{ieee_fullname}
\bibliography{egbib}
}